\pdfoutput=1
\documentclass[11pt]{article}

\usepackage[]{acl}

\usepackage{times}
\usepackage{latexsym}
\usepackage{amsmath,amsfonts,amssymb,amsthm}
\usepackage{comment}
\usepackage{booktabs}

\usepackage{caption}
\usepackage{subcaption}
\usepackage[ruled,noline,noend,lined]{algorithm2e}
\SetKwComment{Comment}{/* }{ */}
\usepackage{multirow}
\usepackage{array}
\usepackage{soul}

\usepackage{colortbl}

\SetAlCapNameFnt{\footnotesize}
\SetAlCapFnt{\footnotesize}

\usepackage{float}
\usepackage{graphicx}
\usepackage[T1]{fontenc}
\usepackage[utf8]{inputenc}

\usepackage{microtype}

\usepackage{tipa}
\newcommand{\switchtosinglerow}{%
  \twocolumn
  \clearpage
  \onecolumn
}

\usepackage{comment}
\usepackage{supertabular}
\usepackage{graphics}
\usepackage{color,soul}
\usepackage{booktabs}
\usepackage{paralist}
\usepackage[noend]{algpseudocode}
\usepackage{booktabs}
\usepackage{hvfloat}
\usepackage{comment}
\usepackage{url}
\usepackage[utf8]{inputenc}
\usepackage{lipsum}
\usepackage{afterpage}
\usepackage{xcolor}
\usepackage{tabularx}
\usepackage{wallpaper}
\usepackage{adjustbox}
\usepackage[normalem]{ulem}
\useunder{\uline}{\ul}{}
\usepackage{rotating}
\usepackage{listings}
\usepackage[english]{babel}
\usepackage{amsmath}
\usepackage{amsfonts}
\usepackage{amssymb}
\usepackage[justification=centering]{caption}
\usepackage{fontenc}
\usepackage{array}
\usepackage{relsize}
\usepackage{subcaption}
\usepackage{caption}
\usepackage{tcolorbox}
\usepackage{lscape}
\usepackage{lastpage}
\usepackage{acro}
\usepackage[toc,page]{appendix}
\usepackage[nottoc]{tocbibind} % for show references in toc
\usepackage{pgfgantt}
\colorlet{punct}{red!60!black}
\definecolor{background}{HTML}{EEEEEE}
\definecolor{delim}{RGB}{20,105,176}
\colorlet{numb}{magenta!60!black}
\lstdefinelanguage{json}{
    basicstyle=\normalfont\ttfamily,
    numbers=left,
    numberstyle=\scriptsize,
    stepnumber=1,
    numbersep=8pt,
    showstringspaces=false,
    breaklines=true,
    frame=lines,
    backgroundcolor=\color{background},
    literate=
     *{0}{{{\color{numb}0}}}{1}
      {1}{{{\color{numb}1}}}{1}
      {2}{{{\color{numb}2}}}{1}
      {3}{{{\color{numb}3}}}{1}
      {4}{{{\color{numb}4}}}{1}
      {5}{{{\color{numb}5}}}{1}
      {6}{{{\color{numb}6}}}{1}
      {7}{{{\color{numb}7}}}{1}
      {8}{{{\color{numb}8}}}{1}
      {9}{{{\color{numb}9}}}{1}
      {:}{{{\color{punct}{:}}}}{1}
      {,}{{{\color{punct}{,}}}}{1}
      {\{}{{{\color{delim}{\{}}}}{1}
      {\}}{{{\color{delim}{\}}}}}{1}
      {[}{{{\color{delim}{[}}}}{1}
      {]}{{{\color{delim}{]}}}}{1},
}

\definecolor{upcorange}{HTML}{FFD33E}
\hypersetup{allcolors=black}

\acsetup{first-style=short}

\usepackage{changepage}
\usepackage{setspace}

\title{The power of Prompts: Evaluating and Mitigating \\ Gender Bias in MT with LLMs}
\author{Aleix Sant, \hspace{0.1cm}Carlos Escolano, \hspace{0.1cm}Audrey Mash, \\
\textbf{Francesca De Luca Fornaciari, \hspace{0.1cm}Maite Melero} \vspace{0.1cm} \\
Barcelona Supercomputing Center (BSC) \\
{\small \tt \{aleix.santsavall,carlos.escolano,audrey.mash} \\
{\small \tt francesca.delucafornaciari,maite.melero\}}{\small \tt @bsc.es}
}

\begin{document}
\maketitle
\begin{abstract}
This paper studies gender bias in machine translation through the lens of Large Language Models (LLMs). Four widely-used test sets are employed to benchmark various base LLMs, comparing their translation quality and gender bias against state-of-the-art Neural Machine Translation (NMT) models for English to Catalan (En → Ca) and English to Spanish (En → Es) translation directions. Our findings reveal pervasive gender bias across all models, with base LLMs exhibiting a higher degree of bias compared to NMT models.

To combat this bias, we explore prompting engineering techniques applied to an instruction-tuned LLM. We identify a prompt structure that significantly reduces gender bias by up to 12\% on the WinoMT evaluation dataset compared to more straightforward prompts. These results significantly reduce the gender bias accuracy gap between LLMs and traditional NMT systems. 

\end{abstract}

\section{Introduction}
\label{sec:intro}
Within the domain of machine translation, gender bias is defined as the tendency of MT systems to produce translations that reflect or perpetuate gender stereotypes, inequalities, or assumptions based on cultural and societal biases \cite{10.1145/230538.230561, savoldi-etal-2021-gender}. Given that the presence of such bias can lead to harmful consequences for certain groups — either in representational (i.e., misrepresentation or underrepresentation of social groups and their identities) or allocational harms (i.e., allocation or withholding of opportunities or resources to certain groups) — \cite{Levesque2011, CiteGender, CiteGender2, article_gender}, it becomes paramount to thoroughly investigate and mitigate its occurrence. Nevertheless, addressing gender bias is a multi-faceted task. 

Gender bias is a pervasive issue in all generative NLP models, and LLMs are no exception to this situation. LLMs have gained significant popularity in recent years and are being used for many NLP tasks, including machine translation. While gender bias in machine translation has been extensively studied for Neural Machine Translation models, little attention has been paid to this type of bias in LLMs. This paper aims to address this gap by examining and trying to mitigate this bias in the translations generated by the LLMs.

The aim of this work is twofold. First, a comprehensive benchmarking process is conducted to compare various base LLMs with some state-of-the-art NMT models. The directions of the translations under study are English $\rightarrow$ Catalan and English $\rightarrow$ Spanish. Distinct popular test sets such as FLoRes-200 \cite{DBLP:journals/corr/abs-2207-04672}, WinoMT \cite{stanovsky-etal-2019-evaluating}, Gold BUG \cite{levy-etal-2021-collecting-large}, and MuST-SHE \cite{bentivogli-etal-2020-gender} are used to assess the translation quality and the gender bias of the models.

Following the benchmarking, an investigation into the effectiveness of prompts in mitigating this bias in LLMs is conducted. The purpose of this research is to determine whether well-designed prompts can serve as a useful strategy in addressing bias. While existing literature has explored various approaches to mitigating this bias in Neural Machine Translation models \cite{DBLP:journals/corr/abs-2012-13176, stafanovics-etal-2020-mitigating, saunders-byrne-2020-reducing}, we specifically focus on the realm of LLMs, probing the role of prompts. In this phase of the study, an instruction-tuned LLM is employed, and several prompt engineering techniques are experimented with, including few-shot \cite{Radford2019LanguageMA, DBLP:conf/icml/ZhaoWFK021, DBLP:journals/jmlr/ChowdheryNDBMRBCSGSSTMRBTSPRDHPBAI23}, context-supplying, and chain of thought \cite{DBLP:journals/corr/abs-2201-11903}.

The relevance of this work lies in several insightful findings. Firstly, we demonstrate that base LLMs tend to lag behind NMT models in terms of translation capabilities and gender-bias scores. Afterwards, through an extensive trial-and-error examination into prompting, we present a prompt that, when applied to an instructed LLM, achieves impressive bias mitigation across gender-bias test sets, resulting in an increase of 12.4 and 11.7 in the respective Catalan and Spanish WinoMT scores. Finally, we study how gender-bias mitigation through prompting impacts LLMs translation performance.

The rest of the paper is organized as follows: Section \ref{sec:related_work} reviews relevant research in the field. Section \ref{sec:methodology} details the methodology, including the datasets, models, and evaluation metrics employed. Section \ref{sec:benchmarking} focuses on the benchmarking, while Section \ref{sec:prompting} explores the investigation into prompting to mitigate gender bias. Section \ref{sec:results} presents the results. Finally, Section \ref{sec:discussion} provides a discussion and Section \ref{sec:conclusions} highlights the conclusions of this work.

\section{Gender Bias Statement} \label{subsec:gender_bias}
As previously stated, gender bias may lead to inequalities and harmful consequences. In the context of machine translation, we easily come up with two different motivations to consider this issue seriously. First, the presence of gender bias may affect the representation of genders in certain communities. On the other hand, the majority of users of a machine translation system may not be proficient in at least one of the languages involved in the translation. Producing incorrect gender translations provides inaccurate information, misleading users who are trying to understand the original text from a translation, or causing them to convey a different meaning when relying on MT engines to communicate.

\begin{figure}[h]
    \centering
    \includegraphics[width=1\linewidth]{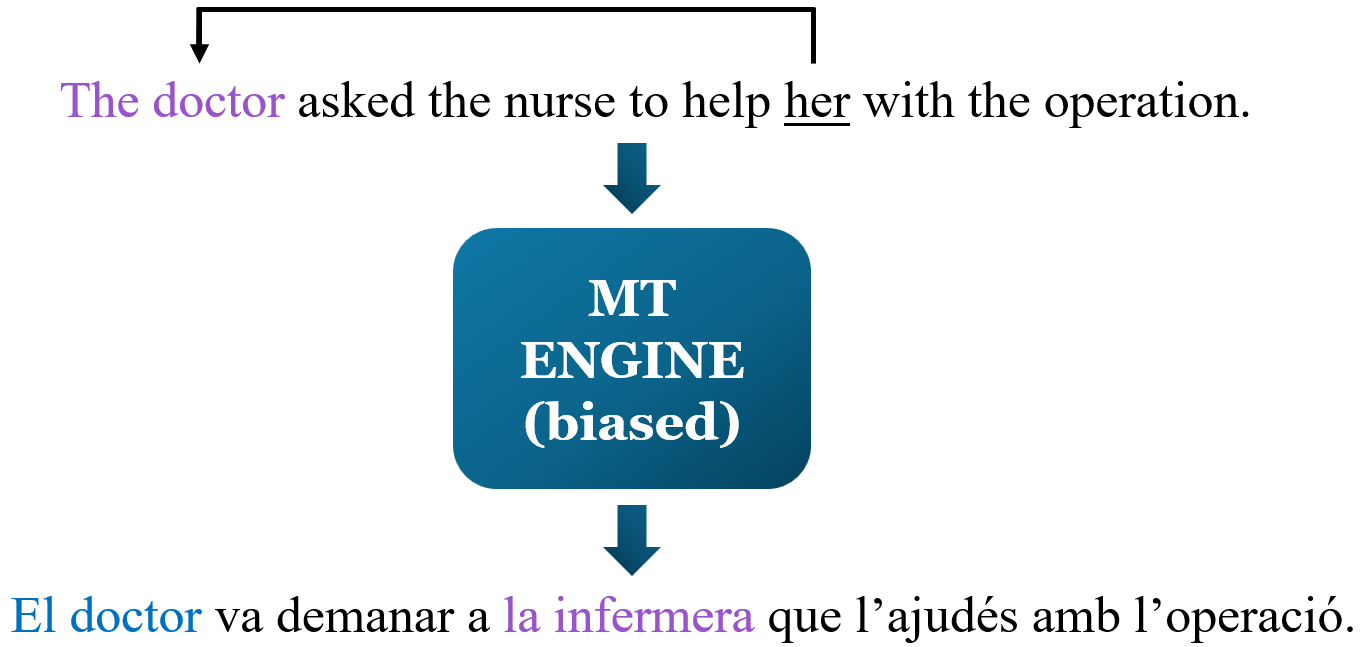}
    \caption{Example of Gender Bias in MT}
    \label{fig:GB_in_MT_example}
\end{figure}

The presence and extent of gender bias in machine translation can vary depending on the languages involved, as gender is manifested differently across languages \cite{CiteBook}. When translating from a language with fewer gender cues to a language with more explicit gender markings, the issue of gender bias can arise. This is precisely the case in our study: we translate from a language with notional gender (English) to languages with grammatical gender (Catalan and Spanish). In this context, certain professions may be stereotypically associated with certain genders. Examples of this phenomenon are \textit{engineers}, who are often translated as masculine, while \textit{nurses} are translated as feminine \cite{CiteForbes}. Additionally, adjectives may be gendered as masculine or feminine based on these stereotypes, rather than relying on gender cues.
Gender pronouns may also be overlooked in favor of or against certain genders. Let's consider a typical example (Figure \ref{fig:GB_in_MT_example}). 

In our study, we analyze gender bias in two distinct ways, which we will refer to as gender-bias tasks: \textit{Gender Coreference Resolution} and \textit{Gender Terms Detection}. In both tasks, models must utilize contextual gender information (i.e., gender cues) to accurately translate, providing the correct gender terms in the translation.  

\paragraph{Gender Coreference Resolution} In this task, we assess whether an MT engine correctly predicts the gender of a human entity in the translation based on its corresponding coreference pronoun in the source sentence. We address this task using POS tagging, focusing solely on the gender of specific human entities in the translation.

\paragraph{Gender Terms Detection} In this other task, we evaluate whether an MT engine generates translations that include all the correct gender terms based on the gender cues of the source sentence. These clues for disambiguating gender terms include coreference pronouns, proper nouns, and semantic meaning, among others. Detection of the correct gender terms (or their incorrect counterparts) relies on textual comparison of reference terms.

\begin{figure*}[h]
    \centering
    \includegraphics[width=1\linewidth]{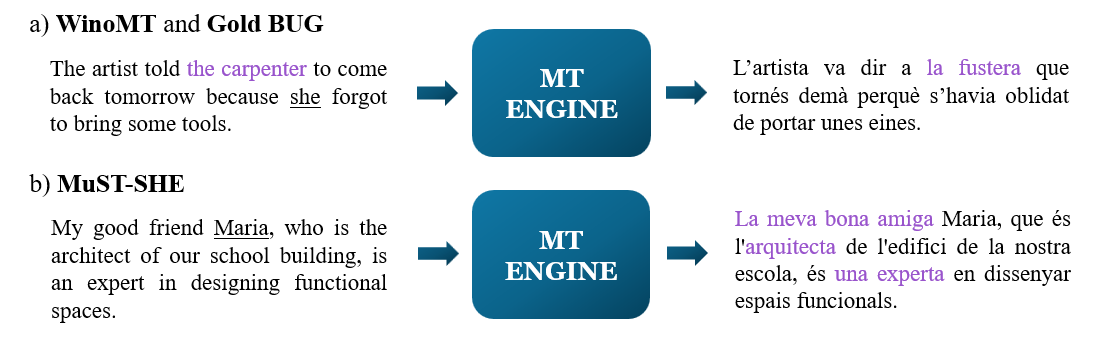}
    \caption{Examples of \textit{Gender Coreference Resolution} (a) and \textit{Gender Terms Detection} (b) in En $\rightarrow$ Ca}
    \label{fig:gb_tasks}
\end{figure*}

\paragraph{}
Both gender-bias problems are approached as classification problems since they involve determining the correct gender labels, allowing for the derivation of typical ML scores. Devoid of gender context, we only pay attention to the proportion of male and female terms generated in the translations. As evaluation benchmarks, WinoMT and Gold BUG focus on \textit{Gender Coreference Resolution}, whereas MuST-SHE in \textit{Gender Terms Detection}. Check Figure \ref{fig:gb_tasks} for an illustration of the gender-bias tasks just explained.

\section{Related work} \label{sec:related_work}
Large Language Models are advanced AI models designed to understand and generate language \cite{DBLP:journals/corr/abs-2304-13712, zhao2023survey}. These models typically employ a decoder-only architecture and are characterized by their enormous size, often containing billions of parameters \cite{DBLP:journals/corr/abs-2005-14165, DBLP:journals/corr/abs-2201-08239, DBLP:journals/corr/abs-2303-08774}. The scale and capacity of LLMs enable them to capture intricate linguistic nuances and handle a wide range of language-related tasks, despite not being explicitly trained for each specific task \cite{DBLP:conf/emnlp/SunL0WGZ023, DBLP:journals/corr/abs-2302-10205, DBLP:journals/corr/abs-2304-11633, DBLP:journals/corr/abs-2303-03836, DBLP:conf/nips/YaoYZS00N23, DBLP:conf/emnlp/YangTPK22, DBLP:conf/emnlp/GaoYYC23, DBLP:journals/corr/abs-2305-12943}. The training process for LLMs typically consists of two steps. First, they undergo self-supervised pretraining using vast amounts of text data, which allows them to develop a general understanding of language (i.e., base LLMs). Subsequently, they are fine-tuned on specific supervised tasks to specialize in various applications \cite{DBLP:journals/corr/abs-2210-11416, DBLP:conf/iclr/SanhWRBSACSRDBX22}. One of the key features of LLMs is the prompting mechanism. A prompt serves as the input or activation signal provided to the model. Through this input, we specify to the model the NLP task we want it to perform, such as translation in our case.

By leveraging the ability to guide the model with prompts, instruction-tuned LLMs are created \cite{DBLP:journals/corr/abs-2308-10792}. These are base LLMs that have undergone additional fine-tuning using datasets of instructions, containing explicit instructions or prompts to enhance their performance on various tasks. Instruction-tuning is a subsequent step that tailors the model’s behavior and output according to specific instructions or guidelines \cite{DBLP:conf/acl/MishraKBH22, DBLP:conf/acl/MuennighoffWSRB23, DBLP:conf/icml/LongpreHVWCTZLZ23}.

Moving away from LLMs, we find Neural Machine Translation models \cite{DBLP:conf/ssst/ChoMBB14, DBLP:journals/corr/BahdanauCB14, DBLP:conf/emnlp/LuongPM15, DBLP:journals/tacl/JohnsonSLKWCTVW17, DBLP:journals/corr/WuSCLNMKCGMKSJL16, DBLP:journals/corr/abs-2010-11125, DBLP:journals/corr/abs-2207-04672}. These models represent the state-of-the-art in machine translation, consistently achieving the highest translation performance. They typically leverage an encoder-decoder transformer \cite{DBLP:conf/nips/VaswaniSPUJGKP17} trained with parallel data in a supervised manner, intended solely for the task of translation. Unlike LLMs, NMT models are relatively smaller in size and present unique challenges when it comes to scaling (e.g., bidirectional processing, the attention mechanism complexity...). However, besides their size, the main distinction between LLMs and NMT models lies in the prompting method. LLMs necessitate a prompt to operate, making them entirely dependent on context. This is precisely the aspect we aim to explore: whether we can use the prompting mechanism, absent in NMT models, to alleviate gender bias.

To date, significant research has been conducted on the translation capabilities of LLMs, as extensively documented in the literature \cite{DBLP:journals/jmlr/ChowdheryNDBMRBCSGSSTMRBTSPRDHPBAI23, DBLP:journals/corr/abs-2301-08745, DBLP:journals/corr/abs-2304-04675, agrawal-etal-2023-context, jiao2023chatgpt, DBLP:conf/icml/0006HB23, DBLP:conf/eamt/BawdenY23, DBLP:journals/corr/abs-2302-09210}. Furthermore, several efforts have been made to identify and address bias in LLMs \cite{DBLP:conf/aequitas/ErnstMBVFKL23, DBLP:journals/corr/abs-2310-11079, DBLP:journals/corr/abs-2403-14409}. However, the exploration of gender bias in the realm of MT and LLMs remains relatively scarce. This encompasses \cite{DBLP:journals/corr/abs-2309-03175}, which sought to leverage LLMs for gender-specific translations, and \cite{DBLP:journals/corr/abs-2401-10016}, which experimented with En $\rightarrow$ It translation in ChatGPT, revealing how GPT models perpetuate biases even when explicitly prompted to provide alternative translations. Additionally, \cite{DBLP:conf/aies/GhoshC23} examines bias between English and languages that exclusively use gender-neutral pronouns, and \cite{savoldi-etal-2024-prompt} demonstrate through extensive manual analysis the potential of GPT-4 to produce gender-neutral translations for En $\rightarrow$ It.

\section{Methodology} \label{sec:methodology}
\subsection{Models}
%Three base decoder-only LLMs have been selected for the benchmarking: 
\paragraph{Llama-2-7B}
A base model that belongs to a family of state-of-the-art LLMs openly released by Meta \cite{DBLP:journals/corr/abs-2307-09288}. This family of models outperforms open-source models on popular benchmarks and has demonstrated high efficacy and safety based on human evaluations. Llama-2-7B was trained on a combination of publicly available data, primarily in English. Catalan and Spanish (among other languages) were also included to a lesser extent. However, any use of the model in languages other than English is explicitly declared out of scope by the developers. 

\paragraph{Ǎguila-7B} 
An open-source base LLM from Barcelona Supercomputing Center (BSC) that was trained on a combination of Spanish, Catalan, and English data, resulting in a total of 26 billion tokens. The model was built upon the Falcon-7B model, which is a highly advanced English language model.

\paragraph{Flor-6.3B}
Another publicly available base LLM tailored for Catalan, Spanish, and English, published by the BSC. This model is derived from the language adaptation process applied to Bloom-7.1B, involving adjustments to the vocabulary and embedding layer. Additionally, the model underwent continuous pre-training with 140 billion tokens specific to Catalan and Spanish. 

\paragraph{M2M-100-1.2B}
A multilingual NMT model released by Meta in October 2020 \cite{DBLP:journals/corr/abs-2010-11125} that can directly translate between the 9,900 directions of 100 languages, including our languages of interest (i.e., English, Catalan, and Spanish). It was considered the first AI model that could translate between 100 languages without relying on English.

\paragraph{NLLB-200-1.3B}
The following multilingual NMT model released by Meta in July 2022 \cite{DBLP:journals/corr/abs-2207-04672} enabling translation across 200 languages, including less commonly spoken languages. It also incorporates the languages we are concentrating on, namely English, Catalan, and Spanish.

\paragraph{Mt-aina-en-ca}
The only parallel NMT model assessed in this work, functioning exclusively for English $\rightarrow$ Catalan translation. Developed at BSC, it was trained from scratch employing a combination of English-Catalan datasets consisting of approximately 11 million sentences.

\paragraph{Google Translate}
It is widely acknowledged in the literature as one of the leading translation models of today. This multilingual NMT model encompasses 133 languages, with English, Catalan and Spanish among them.

\paragraph{Llama-2-7B-chat}
It is the refined iteration of Llama-2-7B, optimized specifically for dialogue applications. This version underwent supervised instruction-tuning as well as Reinforcement Learning from Human Feedback (RLHF). Opting for this instructed version for the investigation into prompting is preferable over the base model, as it is more robust to prompt variations and better comprehends complex prompts and nuances. Selecting the base model along with its instructed version allows us to make insightful comparisons between these models.

\subsection{Test sets}
All test sets comprise English sentences (or paragraphs) aimed to be translated into either Catalan or Spanish. After obtaining translations in their respective grammatical languages, the evaluation frameworks are applied to derive the metrics (either MT or gender scores).

\subsubsection{Machine Translation}
\paragraph{FLoRes-200} It is a massively multilingual general domain dataset. Initially presented by \cite{guzman-etal-2019-flores, DBLP:journals/corr/abs-2106-03193}, it has been further developed and expanded by the \cite{goyal-etal-2022-flores}. The most recent version of this dataset encompasses 200 languages \cite{DBLP:journals/corr/abs-2207-04672}. This dataset\footnote{\url{https://github.com/facebookresearch/flores/tree/main/flores200}} includes two subsets: FLoRes-200 dev (997) and FLoRes-200 devtest (1,012).

\subsubsection{Gender Bias}
\paragraph{WinoMT}
Developed by \cite{stanovsky-etal-2019-evaluating}, this test set is intended to evaluate the presence of gender bias in translations from English to various gender-inflected languages. The corpus\footnote{\url{https://github.com/gabrielStanovsky/mt_gender}} consists of 3,888 sentences in the schema of Winograd. Each sentence in the corpus presents two human entities defined by their roles, along with a subsequent pronoun that must be correctly resolved to one of the entities \cite{DBLP:conf/kr/LevesqueDM12}. One of the main limitations of this dataset is its synthetic nature, as it is built on templates.

\paragraph{Gold BUG}
The previous limitation of WinoMT could be addressed through the introduction of BUG\footnote{\url{https://github.com/SLAB-NLP/BUG}} \cite{levy-etal-2021-collecting-large}, the first publicly accessible large-scale corpus designed for gender-bias evaluation, comprising 108,000 real-world English sentences. BUG was built by crawling text according to specific syntactic patterns, offering a more diverse and realistic dataset than WinoMT. The Gold BUG version used in our evaluation consists of a gold-quality, human-validated set extracted from BUG, totaling 1,717 instances.

\paragraph{MuST-SHE}
This test set, initially introduced by \cite{bentivogli-etal-2020-gender} for English-French, English-Italian, and English-Spanish, serves as a valuable benchmark for evaluating gender bias in the context of speech translation. This dataset\footnote{\url{https://mt.fbk.eu/must-she/}} is constructed using TED talks data, as described by \cite{CATTONI2021101155}, lending it a more natural and realistic tone. Recently, \cite{mash-etal-2024-unmasking-biases} created an English-Catalan\footnote{\url{https://huggingface.co/datasets/projecte-aina/MuST-SHE_en-ca}} version of the dataset tailored for the machine translation domain, resulting in 1,046 sentences. For our analysis, we adapted the original English-Spanish version for machine translation following the same steps as in the Catalan version, resulting in 1,164 instances. Both datasets, English-Spanish and English-Catalan, contain two types of instances: those with and without cues to disambiguate the gender of certain terms. In instances where gender cues are present, the task to be addressed is \textit{Gender Terms Detection}; otherwise, we are solely interested in the proportion of male and female terms generated in the translations.

Furthermore, both WinoMT and Gold BUG contain pro- and anti-stereotypical sets based on US labor statistics \cite{zhao-etal-2018-gender}. A pro-stereotypical set comprises sentences with stereotypical gender-role assignments (e.g., male doctors, female housekeepers), while an anti-stereotypical set includes sentences with non-stereotypical gender-role assignments (e.g., female doctors, male housekeepers). These sets facilitate the investigation of whether the translation performance of models correlates with gender stereotypes. Specifically, they help determine whether models exhibit better or worse gender scores when translating sentences that align (or do not align) with their pre-established biases. %In other words, they aid in determining if the model relies on stereotypes to generate translations.

\subsection{Metrics}
\subsubsection{Machine Translation}
To measure the MT capabilities of the models, we employ two widely-used metrics: BLEU \cite{papineni-etal-2002-bleu}, which is based on comparing n-grams and is computed using the SacreBLEU library\footnote{Version 1.5.1} \cite{post-2018-call}, and COMET \cite{rei-etal-2020-comet}, a more recent metric that relies on sentence embeddings.

\subsubsection{Gender Bias}
When the source sentence contains gender cues to disambiguate the gender of certain terms, meaning we have a known gender reference or ground truth, the translation problem is treated as a typical classification task. Consequently, in the context of gender bias, we evaluate models using Gender Accuracy (in \%), F1-male, and F1-female scores. For WinoMT and Gold BUG these scores are computed directly\footnote{\url{https://github.com/gabrielStanovsky/mt_gender/blob/master/scripts/evaluate_all_languages.sh}}, whereas for MuST-SHE we obtain first the confusion matrix\footnote{\url{https://github.com/audreyvm/tfm_gender_bias/blob/main/mustshe_acc_v1.2.py}} and then we compute the scores using scikit-learn library. Additionally, we get standard metrics such as $\Delta$G, which indicates the performance difference between correctly predicting male and female terms, and $\Delta$S, which requires both a pro- and anti-stereotypical sets to assess whether a model relies on gender-stereotypes to generate translations. Conversely, when no gender cues are available in the source sentence, we simply analyze the proportion of predicted male and female terms in the translations. 

\section{Benchmarking} \label{sec:benchmarking}
\subsection{Prompting LLMs}
In our benchmarking, we employ a 5-shot approach for our LLMs. This ensures that the LLMs better comprehend the requested task (i.e., machine translation) and potentially produce higher-quality translations, as demonstrated in existing literature \cite{DBLP:conf/acl/VilarFCLRF23, DBLP:conf/icml/GarciaBCFKJF23, zhang-etal-2023-machine}. Additionally, during our experimentation with prompts, we observe that incorporating the language label followed by a colon (e.g., ``English:'', ``Catalan:'',``Spanish:'') before the sentence to be translated and its corresponding translation is an effective strategy for our LLMs. Furthermore, \textit{beginning} and \textit{end of sentence} tokens are added to delimit the source and translation examples in the shots, enhancing the models' understanding and facilitating the extraction of the output translations. Flor-6.3B and Llama-2-7B work with ``<BOS>'' and ``<EOS>'', while Ǎguila-7B uses ``<s>'' and ``</s>''.

When evaluating the FLoRes-200 dev set, we use 5 shots from the FLoRes-200 devtest set in the prompt. Conversely, when assessing the FLoRes-200 devtest set, we incorporate 5 shots from the FLoRes-200 dev set into the prompt. For the remaining gender-bias test sets (WinoMT, Gold BUG, and MuST-SHE), we utilize the same prompt employed during testing of the FLoRes-200 dev test set, consisting of the same 5 shots from the FLoRes-200 devtest. When selecting these 5 instances to serve as shots, we ensure diversity in content, length, and structure to provide a broader range of examples to the model. The specific prompts created are detailed in Section \ref{sec:prompts_bench} of the Appendix.

\begin{table*}[t]
  \begin{adjustwidth}{-1cm}{-1cm} 
  \centering
  \renewcommand{\arraystretch}{1.2} 
  \begin{tabular}{|c|c|c|c|c|c|c|c|c|c|}
    \cline{3-10}
    \multicolumn{2}{c|}{} & \multicolumn{4}{|c|}{English $\rightarrow$ Catalan} & \multicolumn{4}{|c|}{English $\rightarrow$ Spanish} \\
    \cline{3-10}
    \multicolumn{2}{c|}{} & \multicolumn{2}{|c|}{\small DEV} & \multicolumn{2}{|c|}{\small DEVTEST} & \multicolumn{2}{|c|}{\small DEV} & \multicolumn{2}{|c|}{\small DEVTEST} \\
    \cline{3-10}
    \cline{3-10} 
        \multicolumn{2}{c|}{} & \small \textbf{BLEU} & \small \textbf{COMET} & \small \textbf{BLEU} & \small \textbf{COMET} & \small \textbf{BLEU} & \small \textbf{COMET} & \small  \textbf{BLEU} & \small \textbf{COMET} \\
    \hline
    \multirow{4}{*}{\rotatebox[origin=c]{90}{NMT}} & \cellcolor[HTML]{C9DAF8} \small Google Translate & \small \textbf{\textcolor{blue}{45.1}} & \small \textbf{\textcolor{blue}{0.8838}} &  \small \textbf{\textcolor{blue}{46.0}} & \small \textbf{\textcolor{blue}{0.8811}} & \small  \textbf{\textcolor{blue}{29.6}} & \small  \textbf{\textcolor{blue}{0.8737}} & \small  \textbf{\textcolor{blue}{30.1}} & \small  \textbf{\textcolor{blue}{0.8724}} \\
    \cline{2-10}
     & \cellcolor[HTML]{C9DAF8} \small NLLB-200-1.3B & \small 38.7 & \small 0.8645 & \small 38.6 & \small 0.8626 & \small 27.2 & \small 0.8591 & \small 27.7 & \small 0.8578 \\
    \cline{2-10}
     & \cellcolor[HTML]{C9DAF8} \small  M2M-100-1.2B &  \small 40.1 & \small 0.8687 & \small 40.4 & \small 0.8623 & \small 25.6 & \small 0.8450 & \small 25.4 & \small 0.8422 \\
    \cline{2-10}
     & \cellcolor[HTML]{C9DAF8} \small Mt-aina-en-ca & \small 43.0 & \small 0.8735 &  \small 43.9 & \small 0.8730 & - & - & - & - \\
    \hline
    \hline
    \multirow{6}{*}{\rotatebox[origin=c]{90}{LLM}}
     & \small  \cellcolor[HTML]{F4CCCC}  Ǎguila-7B & \small 29.1 & \small 0.8359 & \small 30.3 & \small 0.8368 & \small 18.2 & \small 0.8212 & \small 19.5 & \small 0.8198 \\
    \cline{2-10}
    & \small  \cellcolor[HTML]{F4CCCC}  Flor-6.3B & \small \textbf{\textcolor{red}{37.9}} & \small \textbf{\textcolor{red}{0.8641}} & \small \textbf{\textcolor{red}{39.6}} & \small \textbf{\textcolor{red}{0.8680}} & \small \textbf{\textcolor{red}{23.8}} & \small \small \textbf{\textcolor{red}{0.8498}} & \small \textbf{\textcolor{red}{25.5}} & \small \textbf{\textcolor{red}{0.8528}} \\
    \cline{2-10}
     & \small  \cellcolor[HTML]{F4CCCC} Llama-2-7B & \small 31.6 & \small 0.8443 & \small 32.9 & \small 0.8458 & \small 23.3 & \small 0.8486 & \small 23.5 & \small 0.8454  \\
    \cline{2-10}
     & \cellcolor[HTML]{fffcbc} \small Llama-2-7B-chat & \small \textbf{\textcolor[HTML]{ff9e00}{30.1}} & \small \textbf{\textcolor[HTML]{ff9e00}{0.8284}} &  \small \textbf{\textcolor[HTML]{ff9e00}{29.9}} &  \small \textbf{\textcolor[HTML]{ff9e00}{0.8250}} & \small \textbf{\textcolor[HTML]{ff9e00}{22.6}} & \small \textbf{\textcolor[HTML]{ff9e00}{0.8427}} & \small \textbf{\textcolor[HTML]{ff9e00}{22.9}} & \small \textbf{\textcolor[HTML]{ff9e00}{0.8423}}  \\
    \cline{2-10}
    & \cellcolor[HTML]{fffcbc} \small Llama-2-7B-chat &  &   &  & & & & &  \\
    & \cellcolor[HTML]{fffcbc} \small (GB prompt) & \multirow{-2}{*}{ \small 27.6} & \multirow{-2}{*}{\small 0.8176} & \multirow{-2}{*}{\small 28.4} &  \multirow{-2}{*}{\small 0.8140} & \multirow{-2}{*}{\small 22.4} & \multirow{-2}{*}{\small 0.8251} &  \multirow{-2}{*}{\small 21.8} & \multirow{-2}{*}{\small 0.8277}  \\
    \hline
  \end{tabular}
  \caption{BLEU and COMET scores for FLoRes-200}
  \label{tab:mt}
 \end{adjustwidth}
\end{table*}

\subsection{Configurations}
Since we are only performing inference, we adjust only two parameters: the \textit{top\_k}, which is set to 1 to ensure a deterministic process, and the limit of \textit{maximum tokens} to generate, which is adjusted depending on the test sets. We use greedy decoding for all models since beam search in LLMs demands significant time and resources. These choices are made to ensure the comparability of the results.

\subsection{Key takeaways}
Based on the benchmarking evaluation, the following findings emerge:

\begin{itemize}
    \item Base LLMs fall behind NMT models in terms of MT in both En $\rightarrow$ Ca and En $\rightarrow$ Es directions (check Table \ref{tab:mt} to see the results).
    \item All models exhibit gender bias in the assessed directions, with LLMs showing a more pronounced bias compared to NMT models (check Tables \ref{tab:winomt}, \ref{tab:goldbug}, and \ref{tab:mustshe}).
    \item The performance of all studied models correlates with gender stereotypes, achieving better gender metrics for the pro-stereotypical set rather than the anti-stereotypical set (check Section \ref{sec:pro_and_anti} in the Appendices).
    \item In the absence of contextual gender cues, all models predict mostly male terms ($\sim$75\%-94\%). The corresponding ($\sim$6\%-25\%) mainly relates to female-stereotypical examples (check Section \ref{sec:no_context} in the Appendices). 
\end{itemize}

\section{Gender Bias mitigation through prompting} \label{sec:prompting}
After observing that LLMs exhibit more gender bias than NMT models, we found it necessary to address this bias in LLMs. Consequently, we have chosen to leverage prompting, as it is a distinctive feature of these models. Therefore, the second stage of our work involves conducting exploratory research in a trial-and-error manner, aiming to identify a prompt that effectively mitigates bias in LLMs. For this experiment, we have selected the instruction-tuned model Llama-2-7B-chat since it is more robust to complex prompts than its base version. In addition, in this stage, we have decided to focus solely on the \textit{Gender Coreference Resolution} task. Ideally, our goal is to narrow the gap in gender scores with respect to NMT models, as this would represent a significant breakthrough.

The procedure goes as follows: Initially, we develop a range of prompts based on strategies outlined in the literature, including few-shot prompting, context-supplying, and chain-of-thought instructions. To assess the impact of these prompts, we test them on WinoMT and obtain gender-bias scores for each prompt. Thereafter, the prompt that demonstrates the most considerable reduction in bias on WinoMT, as indicated by numerical gender-bias scores, is evaluated on the remaining test sets (Gold BUG, MuST-SHE, and FLoRes-200). By doing so, we want to determine: firstly, the prompt's generalizability across the remaining gender-bias test sets, and secondly, if it affects the overall machine translation capabilities of the LLM.

\begin{table*}[t]
\resizebox{\textwidth}{!}{
\centering
\renewcommand{\arraystretch}{1.8} 
\begin{tabular}{cp{4cm}|cccc|cccc|}
\cline{3-10}
\multicolumn{1}{l}{}                                   & \multicolumn{1}{l|}{}                                  & \multicolumn{4}{c|}{ \Large{English $\rightarrow$ Catalan}}                                                                                                       & \multicolumn{4}{c|}{\Large{English $\rightarrow$ Spanish}}                                                                                                                            \\ \hline
\multicolumn{1}{|c|}{\large \textbf{Model}}                            & \large \textbf{Examples from:}                                         & \multicolumn{1}{c|}{\textbf{G Acc}} & \multicolumn{1}{c|}{\textbf{F1-male}} & \multicolumn{1}{c|}{\textbf{F1-female}} & \multicolumn{1}{l|}{\textbf{$\Delta$G}} & \multicolumn{1}{c|}{\textbf{G Acc}} & \multicolumn{1}{c|}{\textbf{F1-male}} & \multicolumn{1}{l|}{\textbf{F1-female}} & \multicolumn{1}{l|}{\textbf{$\Delta$G}} \\ \hline
\multicolumn{1}{|c|}{Llama-2-7B}                       & FLoRes-200                                             & \multicolumn{1}{c|}{48.0}           & \multicolumn{1}{c|}{62.6}             & \multicolumn{1}{c|}{36.4}               & 26.2                             & \multicolumn{1}{c|}{53.1}           & \multicolumn{1}{c|}{64.9}             & \multicolumn{1}{c|}{41.3}               & 23.6                                                  \\ \hline \hline
\multicolumn{1}{|c|}{\multirow{10}{*}{Llama-2-7B-chat}} & FLoRes-200                                             & \multicolumn{1}{c|}{46.4}           & \multicolumn{1}{c|}{61.4}             & \multicolumn{1}{c|}{35.5}               & 25.9                             & \multicolumn{1}{c|}{53.3}           & \multicolumn{1}{c|}{65.3}             & \multicolumn{1}{c|}{41.6}               & 23.7                                                  \\ \cline{2-10} 
\multicolumn{1}{|c|}{}                                 & MuST-SHE                                               & \multicolumn{1}{c|}{46.9}           & \multicolumn{1}{c|}{60.6}             & \multicolumn{1}{c|}{39.4}               & 21.2                             & \multicolumn{1}{c|}{49.9}           & \multicolumn{1}{c|}{62.7}             & \multicolumn{1}{c|}{35.4}               & 27.3                                                  \\ \cline{2-10} 
\multicolumn{1}{|c|}{}                                 & Invented Winograd examples                                & \multicolumn{1}{c|}{46.6}           & \multicolumn{1}{c|}{58.8}             & \multicolumn{1}{c|}{43.2}               & 15.6                             & \multicolumn{1}{c|}{49.8}           & \multicolumn{1}{c|}{61.8}             & \multicolumn{1}{c|}{37.5}               & 24.3                                                  \\ \cline{2-10} 
\multicolumn{1}{|c|}{}                                 & MuST-SHE + context on Gender Bias issue                & \multicolumn{1}{c|}{46.9}           & \multicolumn{1}{c|}{60.0}             & \multicolumn{1}{c|}{40.7}               & 19.3                             & \multicolumn{1}{c|}{50.6}           & \multicolumn{1}{c|}{62.6}             & \multicolumn{1}{c|}{38.7}               & 23.9                                                  \\ \cline{2-10} 
\multicolumn{1}{|c|}{}                                 & Invented Winograd examples + chain-of-thought ("agent") & \multicolumn{1}{c|}{55.2}           & \multicolumn{1}{c|}{65.8}             & \multicolumn{1}{c|}{56.7}               & 9.1                              & \multicolumn{1}{c|}{60.6}           & \multicolumn{1}{c|}{69.7}             & \multicolumn{1}{c|}{56.8}               & 12.9                                                  \\ \cline{2-10} 
\multicolumn{1}{|c|}{}                                 & Invented Winograd examples + chain-of-thought ("human entity") & \multicolumn{1}{c|}{54.5}           & \multicolumn{1}{c|}{65.2}             & \multicolumn{1}{c|}{55.3}               & 9.9                              & \multicolumn{1}{c|}{59.5}           & \multicolumn{1}{c|}{68.9}             & \multicolumn{1}{c|}{54.8}               & 14.1                                                  \\ \cline{2-10} 
\multicolumn{1}{|c|}{}                                 & Invented Winograd examples + SHORT chain-of-thought           & \multicolumn{1}{c|}{\textbf{58.8}}  & \multicolumn{1}{c|}{\textbf{68.9}}    & \multicolumn{1}{c|}{\textbf{60.5}}      & \textbf{8.4}                     & \multicolumn{1}{c|}{\textbf{65.0}}  & \multicolumn{1}{c|}{\textbf{73.3}}    & \multicolumn{1}{c|}{\textbf{63.6}}      & \textbf{9.7}                                          \\ \hline
\end{tabular}}
\caption{WinoMT scores using different prompting techniques for En $\rightarrow$ Ca and En $\rightarrow$ Es}
\label{tab:winomt_prompts}
\end{table*}

\subsection{Baseline}
Before embarking on the search for the prompt, it is essential to establish a baseline for Llama-2-7B-chat. Therefore, we use the same prompt employed in the benchmarking, with minor adaptations necessary for Llama-2-7B-chat, such as the use of special tags (<<SYS>>, [INST]...). The resulting prompt after the adjustments is detailed in Section \ref{sec:prompts_llama2chat} of the Appendices. With this prompt, we obtain MT and gender-bias scores across the four test sets. Refer to Tables \ref{tab:mt}, \ref{tab:winomt}, \ref{tab:goldbug}, and \ref{tab:mustshe} to observe the results. These initial results offer valuable insights, revealing that the instructed version (Llama-2-7B-chat) achieves lower MT scores compared to its base model (Llama-2-7B) for both directions. %En $\rightarrow$ Ca and En $\rightarrow$ Es.

\subsection{Crafting and testing prompts on WinoMT}
After conducting several experiments using Llama-2-7B-chat, we proceeded to curate and test multiple prompts on the WinoMT test set. The curated prompts in detail can be found in section \ref{sec:prompts_inv} of the Appendices. We recommend consulting them for a comprehensive understanding of this section.

In the design of all our prompts, we incorporated the 5-shot strategy already used in the benchmarking and the baseline. However, we substituted the FLoRes-200 examples and introduced additional modifications to the curated prompts.

A significant aspect of our crafted prompts involves the inclusion of translation examples that encompass more gender-related phenomena compared to the ones from the FLoRes-200 dataset, which comprises mainly gender-neutral or impersonal sentences. Specifically, one of our curated prompts included examples from the MuST-SHE dataset, while in another prompt, we intentionally created five sentences (or rather, translations) adhering to a Winograd structure, wherein each sentence comprises two human entities and one pronoun used to disambiguate one of them. These crafted translations were deliberately designed to contain more female representation and anti-stereotypical content. These invented translations are provided in section \ref{sec:sentences} of the Appendices.

For another prompt, in addition to including 5 shots from MuST-SHE, we also adopted an approach that involved providing more contextual information to the model. We explicitly stated the objective of translating while simultaneously reducing gender bias. By offering this additional context, the model should gain a clearer understanding of the goal to mitigate gender bias and the factors it should consider to do so effectively.

Afterwards, we adopted a chain-of-thought strategy for the remaining curated prompts, each following again a 5-shot structure. We integrated the previously crafted Winograd examples into these prompts. Two of them resulted in complex and detailed chain-of-thought prompts, incorporating all the necessary steps and reasoning that the model should do to carefully solve the \textit{Gender Coreference Task} and provide a correct translation. The only distinction between these two complex prompts was the terminology used to refer to the human entities in the examples, either as ``human entity'' or ``agent''.

Finally, we constructed another chain-of-thought prompt that 
yielded the best results. In this prompt, the steps were significantly simplified compared to the previous two prompts. Here, explicit instructions of the steps were not included, and instead, schematic steps accompanied by arrows were provided in the shots.

For a comprehensive summary of the results obtained on WinoMT for all these prompts, please consult Table \ref{tab:winomt_prompts}.

\begin{table*}[t]
\resizebox{\textwidth}{!}{
  \centering
  \renewcommand{\arraystretch}{1.2} 
  \begin{tabular}{|c|c|c|c|c|c|c|c|c|c|c|c|}
    \cline{3-12}
    \multicolumn{2}{c|}{} & \multicolumn{5}{|c|}{English $\rightarrow$ Catalan} & \multicolumn{5}{|c|}{English $\rightarrow$ Spanish} \\
    \cline{3-12}
    \multicolumn{2}{c|}{} & \small \textbf{G Acc} & \small \textbf{F1-male} & \small \textbf{F1-female} & \small \textbf{$\Delta$G} & \small \textbf{$\Delta$S} & \small \textbf{G Acc} & \small \textbf{F1-male} & \small \textbf{F1-female} & \small \textbf{$\Delta$G} & \small \textbf{$\Delta$S} \\
    \hline
    \multirow{4}{*}{\rotatebox[origin=c]{90}{NMT}} & \cellcolor[HTML]{C9DAF8} \small Google Translate & \small 57.1 & \small 67.5 & \small 55.6 & \small 11.9 & \small \textbf{\textcolor{blue}{23.9}} & \small \textbf{\textcolor{blue}{70.9}} &   \small \textbf{\textcolor{blue}{76.6}}& \small\textbf{\textcolor{blue}{74.4}} &   \small \textbf{\textcolor{blue}{2.2}} & \small \textbf{\textcolor{blue}{24.3}} \\
    \cline{2-12}
     & \cellcolor[HTML]{C9DAF8} \small NLLB-200-1.3B & \small \textbf{\textcolor{blue}{60.9}} & \small  \textbf{\textcolor{blue}{70.1}} & \small \textbf{\textcolor{blue}{64.0}}  & \small \textbf{\textcolor{blue}{6.1}} & \small 28.1 & \small 67.2 &   \small 74.0 & \small 68.9 & \small 5.1 & \small 33.9\\
    \cline{2-12}
     & \cellcolor[HTML]{C9DAF8} \small M2M-100-1.2B  & \small 51.5 & \small 64.2 & \small 44.6 & \small 19.6 & \small 24.6 & \small 57.9 &  \small 68.6 &  \small 50.4 &  \small 18.2 & \small 26.5  \\
    \cline{2-12}
     & \cellcolor[HTML]{C9DAF8} \small Mt-aina-en-ca & \small 48.9 & \small 63.1 & \small 37.9 & \small 25.2 & \small 27.3 & - & - & - & - & - \\
    \hline
    \hline
    \multirow{6}{*}{\rotatebox[origin=c]{90}{LLM}}
     &   \cellcolor[HTML]{F4CCCC}  \small Ǎguila-7B &  \small  46.1 &  \small  60.4 &   \small 34.5 &    \small \textbf{\textcolor{red}{25.9}} &   \small 36.1 &  \small 49.3 & \small  63.3 &  \small 32.5 & \small  30.8 & \small  \textbf{\textcolor{red}{28.4}} \\
    \cline{2-12}
    &  \cellcolor[HTML]{F4CCCC} \small Flor-6.3B &  \small  47.7 &  \small  62.2 &   \small 35.2 & \small   27.0 &  \small  33.1 & \small \textbf{\textcolor{red}{53.4}} &   \small \textbf{\textcolor{red}{65.1}} &   \small\textbf{\textcolor{red}{42.5}} & \small  \textbf{\textcolor{red}{22.6}} & \small  30.1 \\
    \cline{2-12}
     &  \cellcolor[HTML]{F4CCCC} \small Llama-2-7B &   \small \textbf{\textcolor{red}{48.0}} &   \small \textbf{\textcolor{red}{62.6}} &  \small  \textbf{\textcolor{red}{36.4}} &  \small  26.2 & \small   \textbf{\textcolor{red}{32.8}} & \small 53.1 & \small  64.9 &  \small 41.3 &  \small 23.6 & \small  33.1 \\
    \cline{2-12}
     &  \cellcolor[HTML]{fffcbc} \small Llama-2-7B-chat &  \small 46.4 &  \small 61.4 &  \small  35.5 &  \small 25.9 & \small   33.1 & \small 53.3 & \small 65.3 & \small 41.6 & \small 23.7 &  \small 32.0 \\
     \cline{2-12}
     &  \cellcolor[HTML]{fffcbc} \small Llama-2-7B-chat &   &   &   &  &  &  &  &  &  &  \\
     &  \cellcolor[HTML]{fffcbc} \small (GB prompt) &  \small \multirow{-2}{*}{\textbf{\textcolor[HTML]{ff9e00}{58.8}}} & \small \multirow{-2}{*}{\textbf{\textcolor[HTML]{ff9e00}{68.9}}} &  \small  \multirow{-2}{*}{\textbf{\textcolor[HTML]{ff9e00}{60.5}}} &  \small \multirow{-2}{*}{\textbf{\textcolor[HTML]{ff9e00}{8.4}}} & \small \multirow{-2}{*}{\textbf{\textcolor[HTML]{ff9e00}{27.8}}} & \small \multirow{-2}{*}{\textbf{\textcolor[HTML]{ff9e00}{65.0}}} & \small \multirow{-2}{*}{\textbf{\textcolor[HTML]{ff9e00}{73.3}}} & \small \multirow{-2}{*}{\textbf{\textcolor[HTML]{ff9e00}{63.6}}} & \small \multirow{-2}{*}{\textbf{\textcolor[HTML]{ff9e00}{9.7}}} & \small \multirow{-2}{*}{\textbf{\textcolor[HTML]{ff9e00}{22.1}}} \\
    \hline
  \end{tabular}}
  \caption{WinoMT gender scores}
  \label{tab:winomt}
\end{table*}

\subsection{Top-performing prompt}
The resulting top-performing prompt on WinoMT is the one named \textit{Invented Winograd examples + SHORT chain-of-thought} from Table \ref{tab:winomt_prompts}. With this prompt, we have achieved remarkable increases of 12.4 (En $\rightarrow$ Ca) and 11.7 (En $\rightarrow$ Es) on WinoMT compared to the baseline. In short, this prompt follows a simplified chain-of-thought approach with 5-shots on anti-stereotypical content and increased female representation. The examples in the prompt were invented following the Winograd sentence structure, designed to address gender coreference. 

The phrase ``Proceed step by step'' is also included before the shots. In the initial experiments, we observed that incorporating this sentence led to the model providing a more structured response. Based on this observation, we replicated the same pattern generated by the LLM in our crafted shots.

\section{Results} \label{sec:results}
After testing our top-performing prompt on the remaining gender-bias test sets, Gold BUG and MuST-SHE, we observe a significant reduction in gender bias within those test sets too. These results are detailed in sections \ref{sec:goldbug_results} and \ref{sec:mustshe_results} of the Appendices. Subsequently, all the three Tables \ref{tab:winomt}, \ref{tab:goldbug}, and \ref{tab:mustshe} demonstrate a remarkable improvement in gender-bias scores, significantly reducing the upper bound in each test set compared to the best NMT model. This places the LLM on par with NMT models in terms of gender bias manifestation. For example, on the WinoMT test set, the model achieves the second-best position in En $\rightarrow$ Ca and the third-best position in En $\rightarrow$ Es. In MuST-SHE, the mitigation is less pronounced as this test set also encompasses other gender-related tasks, unlike WinoMT and Gold BUG, which focus solely on \textit{Gender Coreference Resolution}.

Regarding the MT metrics, we observe a small loss compared to the baseline when testing on FLoRes-200 (Table \ref{tab:mt}).

\section{Discussion} \label{sec:discussion}
Initially, we believed that reducing gender bias through prompting would possibly be straightforward. However, it was surprising to find that the model only began effectively mitigating the bias after implementing the chain-of-thought approach. In fact, the results presented in Table \ref{tab:winomt_prompts} demonstrate that without the chain-of-thought approach and relying solely on the same invented Winograd examples from the top-performing prompt, no improvement was observed. Furthermore, we noticed that describing the problem of gender bias or including MuST-SHE examples did not lead to any improvement. Additionally, we observed that the Llama-2-7B-chat model comprehends and responds better to schematic chain-of-thought prompts compared to highly detailed and elaborate prompts, resulting in higher gender scores in the former case. Besides, the inclusion of the phrase ``Proceed step by step'' seems to be beneficial.

Fortunately, after identifying our successful prompt, we can confidently affirm that leveraging prompting can indeed serve as an effective method to mitigate gender bias in an instructed LLM (at least, for \textit{Gender Coreference Resolution}). %However, this comes at the expense of a slight loss in machine translation capabilities.

\section{Conclusions} \label{sec:conclusions}
This work investigates gender bias in the translation outputs generated by various LLMs through two distinct approaches. 
Firstly, by benchmarking three base models (Ǎguila-7B, Flor-6.3B and Llama-2-7B) using different gender-bias test sets and comparing the results with state-of-the-art NMT models (M2M-100-1.2B, NLLB-200-1.3B, Mt-aina-en-ca, and Google Translate). 
Secondly, by experimenting with the prompting mechanism of an instruction-tuned LLM (Llama-2-7B-chat) and trying to mitigate its gender bias in the output. This study is done in the En $\rightarrow$ Ca and En $\rightarrow$ Es directions. \\ \\
\indent Results reveal the presence of gender bias across all models, with base LLMs exhibiting more gender bias than NMT models. Moreover, the performance of all models correlates with gender stereotypes. In the absence of gender cues in the source sentence, they tend to generate predominantly male terms, while female terms are generated primarily when encountering female-stereotypical content. To mitigate this bias, prompting engineering techniques have been implemented in an instruction-tuned LLM. After curating and testing several prompts, one prompt was identified that resulted in a significant reduction in gender bias, achieving impressive gender scores. The prompt follows a simplified chain-of-thought approach with 5-shots relying on anti-stereotypical content and increased female representation. This prompt enables the instructed LLM to perform competitively in terms of gender scores, achieving results comparable to NMT models and even surpassing some of them. However, it is observed that using this prompt leads to a slight loss in the translation quality.

\section{Ethical statement} \label{sec:ethical_statement}
In this evaluation, we have only focused on the binary male and female genders, without considering other gender identities. Additional experiments on new datasets would be required to assess the performance of these methods on non-binary scenarios. 

About the proposed definition of gender bias, we tried to characterize different aspects of the problem. Even though we recognize that it is a complex problem and our metrics and experiments focus only on some specific manifestations.

% Acknowledgments
\section{Acknowledgments} 
This research has been promoted and financed by the Government of Catalonia through the Aina project, by the Ministerio para la Transformación Digital y de la Función Pública and Plan de Recuperación, Transformación y Resiliencia (Funded by EU – NextGenerationEU within the framework of the project ILENIA with reference 2022/TL22/00215337, 2022/TL22/00215336, 2022/TL22/00215335, 2022/TL22/00215334). It has also been supported by the Horizon Europe program [Grant Number 101135916] and by DeepR3 (TED2021-130295B-C32) (Funded by MCIN/AEI/10.13039/501100011033 and European Union NextGeneration EU/PRTR).

\bibliography{bibliography}
\bibliographystyle{acl_natbib}

%%% ANNEX %%%
\clearpage
\newpage

% Switch to single column format for the appendix
\switchtosinglerow

\begin{appendices} \label{sec:appendices}
\section{Gender Scores on Gold BUG}
\label{sec:goldbug_results}

\begin{table*}[h]
\resizebox{\textwidth}{!}{
  \centering
  \renewcommand{\arraystretch}{1.2} 
  \begin{tabular}{|c|c|c|c|c|c|c|c|c|c|c|c|}
    \cline{3-12}
    \multicolumn{2}{c|}{} & \multicolumn{5}{|c|}{English $\rightarrow$ Catalan} & \multicolumn{5}{|c|}{English $\rightarrow$ Spanish} \\
    \cline{3-12}
    \multicolumn{2}{c|}{} & \small \textbf{G Acc} & \small \textbf{F1-male} & \small \textbf{F1-female} & \small \textbf{$\Delta$G} & \small\textbf{$\Delta$S} & \small \textbf{G Acc} & \small \textbf{F1-male} & \small \textbf{F1-female} & \small \textbf{$\Delta$G} & \small\textbf{$\Delta$S} \\
    \hline
    \multirow{4}{*}{\rotatebox[origin=c]{90}{NMT}} & \cellcolor[HTML]{C9DAF8} \small Google Translate &  \small \textbf{\textcolor{blue}{62.3}} & \small  \textbf{\textcolor{blue}{77.5}} & \small  56.9 &  \small 20.6 &  \small 26.7 & \small 47.5 &  \small 62.8 &  \small 55.0 &  \small \textbf{\textcolor{blue}{7.8}} &  \small \textbf{\textcolor{blue}{14.9}} \\
    \cline{2-12}
     & \cellcolor[HTML]{C9DAF8} \small NLLB-200-1.3B &   \small 62.1 & \small  77.4 &   \small \textbf{\textcolor{blue}{57.9}} &  \small \textbf{\textcolor{blue}{19.5}} &  \small \textbf{\textcolor{blue}{13.6}} & \small \textbf{\textcolor{blue}{65.2}} &  \small \textbf{\textcolor{blue}{79.4}} & \small  \textbf{\textcolor{blue}{61.9}} & \small  17.5 & \small  20.4  \\
    \cline{2-12}
     & \cellcolor[HTML]{C9DAF8} \small  M2M-100-1.2B & \small  60.4 &  \small 76.3 & \small  49.8 &   \small 26.5 &  \small 23.9 & \small 63.8 &  \small 78.5 & \small  56.5 &   \small 22.0 & \small  22.7\\
    \cline{2-12}
     & \cellcolor[HTML]{C9DAF8} \small Mt-aina-en-ca &  \small 60.3 & \small  76.4 & \small  51.2 &   \small 25.2 &   \small 20.5 & - & - & - & - & \\
    \hline
    \hline
    \multirow{6}{*}{\rotatebox[origin=c]{90}{LLM}}
     &  \cellcolor[HTML]{F4CCCC} \small Ǎguila-7B &  \small 54.5 & \small  71.7 &  \small \textbf{\textcolor{red}{43.9}} & \small \textbf{\textcolor{red}{27.8}} & \small  22.5 & \small 58.8 & \small 75.2 & \small  \textbf{\textcolor{red}{46.5}} & \small \textbf{\textcolor{red}{28.7}} & \small 18.8  \\
    \cline{2-12}
    &  \cellcolor[HTML]{F4CCCC}  \small  Flor-6.3B & \small  \textbf{\textcolor{red}{57.8}} & \small  74.5 &  \small 43.0 & \small  31.5 & \small  \textbf{\textcolor{red}{18.6}} & \small \textbf{\textcolor{red}{61.2}} &  \small \textbf{\textcolor{red}{77.1}} &  \small 46.2 & \small  30.9 &  \small \textbf{\textcolor{red}{14.9}} \\
    \cline{2-12}
     &  \cellcolor[HTML]{F4CCCC}  \small Llama-2-7B & \small  57.7 & \small  \textbf{\textcolor{red}{74.9}} &  \small 37.1 & \small  37.8 & \small  18.7 & \small 60.2 &   \small 76.9 &  \small 37.4 &  \small 39.5 & \small  16.1  \\
     \cline{2-12}
     & \cellcolor[HTML]{fffcbc} \small Llama-2-7B-chat & \small 57.8  & \small 74.5 &  \small 39.3 & \small 35.2 & \small 25.3  & \small 58.9  &   \small 75.6 &  \small 37.0 &  \small 38.6 & \small 16.9   \\
     \cline{2-12}
     &  \cellcolor[HTML]{fffcbc} \small Llama-2-7B-chat & \small  & \small &  \small  & \small & \small  & \small   &   \small  &  \small  &  \small  & \small    \\
     & \cellcolor[HTML]{fffcbc} \small (GB prompt) & \small \multirow{-2}{*}{\textbf{\textcolor[HTML]{ff9e00}{59.8}}} & \small \multirow{-2}{*}{\textbf{\textcolor[HTML]{ff9e00}{75.0}}} &  \small \multirow{-2}{*}{\textbf{\textcolor[HTML]{ff9e00}{58.7}}} & \small \multirow{-2}{*}{\textbf{\textcolor[HTML]{ff9e00}{16.3}}} & \small \multirow{-2}{*}{\textbf{\textcolor[HTML]{ff9e00}{15.4}}}  & \small \multirow{-2}{*}{\textbf{\textcolor[HTML]{ff9e00}{63.7}}}  &   \small \multirow{-2}{*}{\textbf{\textcolor[HTML]{ff9e00}{78.5}}}  &  \small \multirow{-2}{*}{\textbf{\textcolor[HTML]{ff9e00}{58.9}}} &  \small \multirow{-2}{*}{\textbf{\textcolor[HTML]{ff9e00}{19.6}}} & \small \multirow{-2}{*}{\textbf{\textcolor[HTML]{ff9e00}{18.1}}}   \\
    \hline 
  \end{tabular}}
  \caption{Gold BUG gender scores}
  \label{tab:goldbug}
\end{table*}

\section{Gender Scores on MuST-SHE}
\label{sec:mustshe_results}

\begin{table*}[h]
 \begin{adjustwidth}{-1cm}{-1cm}
  \centering
  \renewcommand{\arraystretch}{1.2} 
  \begin{tabular}{|c|c|c|c|c|c|c|c|c|c|}
    \cline{3-10}
    \multicolumn{2}{c|}{} & \multicolumn{4}{|c|}{English $\rightarrow$ Catalan} & \multicolumn{4}{|c|}{English $\rightarrow$ Spanish} \\
    \cline{3-10}
    \multicolumn{2}{c|}{} & \small \textbf{G Acc} & \small \textbf{F1-male} & \small \textbf{F1-female} & \small \textbf{$\Delta$G} & \small \textbf{G Acc} & \small \textbf{F1-male} & \small \textbf{F1-female} & \small \textbf{$\Delta$G} \\
    \hline
    \multirow{4}{*}{\rotatebox[origin=c]{90}{NMT}} & \cellcolor[HTML]{C9DAF8} \small Google Translate &   \small 89.5 &  \small 90.6 &  \small 88.0 & \small 2.6 & \small 95.1 & \small 95.5 & \small 94.7 & \small 0.8 \\
    \cline{2-10}
     & \cellcolor[HTML]{C9DAF8} \small NLLB-200-1.3B & \small  \textbf{\textcolor{blue}{93.3}} &   \small \textbf{\textcolor{blue}{93.7}} &  \small \textbf{\textcolor{blue}{92.7}} &   \small \textbf{\textcolor{blue}{1.0}} & \small \textbf{\textcolor{blue}{96.0}} & \small \textbf{\textcolor{blue}{96.2}} & \small \textbf{\textcolor{blue}{95.8}} & \small \textbf{\textcolor{blue}{0.5}}\\
    \cline{2-10}
     & \cellcolor[HTML]{C9DAF8} \small M2M-100-1.2B & \small  84.4 & \small  86.6 & \small 81.4 &  \small 5.2 & \small 87.4 & \small 89.2 & \small 84.8 & \small 4.3 \\
    \cline{2-10}
     & \cellcolor[HTML]{C9DAF8} \small Mt-aina-en-ca &  \small 87.1 &  \small 88.5 &  \small 85.4 & \small  3.1 & - & - & - & -  \\
    \hline
    \hline
    \multirow{6}{*}{\rotatebox[origin=c]{90}{LLM}}
     &  \cellcolor[HTML]{F4CCCC}  \small  Ǎguila-7B &  \small 87.1 &  \small 88.5 & \small  85.4 & \small  3.1  & \small 92.2 &  \small 93.0 &  \small 91.0 &  \small 2.0  \\
    \cline{2-10}
    &  \cellcolor[HTML]{F4CCCC} \small  Flor-6.3B &  \small 89.6 &  \small 90.7 & \small  88.2 & \small  2.5 & \small  93.3 &  \small 93.9 &  \small 92.4 & \small  1.5 \\
    \cline{2-10}
     &  \cellcolor[HTML]{F4CCCC} \small Llama-2-7B &  \small \textbf{\textcolor{red}{91.1}} &   \small \textbf{\textcolor{red}{91.9}} &  \small \textbf{\textcolor{red}{90.0}} &   \small \textbf{\textcolor{red}{1.8}} & \small \textbf{\textcolor{red}{95.1}} &   \small \textbf{\textcolor{red}{94.5}} &  \small \textbf{\textcolor{red}{93.2}} & \small \textbf{\textcolor{red}{1.3}}\\
     \cline{2-10}
     & \cellcolor[HTML]{fffcbc} \small Llama-2-7B-chat & \small 88.1 & \small 89.7 & \small 86.0  & \small 3.7  & \small 91.0 & \small 92.0   &  \small 89.6  &  \small 2.4 \\
     \cline{2-10}
     & \cellcolor[HTML]{fffcbc} \small Llama-2-7B-chat & & &  &  &  &   &    &  \\
     & \cellcolor[HTML]{fffcbc} \small  (GB prompt) & \small \multirow{-2}{*}{\textbf{\textcolor[HTML]{ff9e00}{88.4}}} & \small \multirow{-2}{*}{\textbf{\textcolor[HTML]{ff9e00}{89.8}}} & \small \multirow{-2}{*}{\textbf{\textcolor[HTML]{ff9e00}{86.5}}}  & \small \multirow{-2}{*}{\textbf{\textcolor[HTML]{ff9e00}{3.3}}}  & \small \multirow{-2}{*}{\textbf{\textcolor[HTML]{ff9e00}{92.0}}} & \small \multirow{-2}{*}{\textbf{\textcolor[HTML]{ff9e00}{92.6}}}  & \small \multirow{-2}{*}{\textbf{\textcolor[HTML]{ff9e00}{91.4}}}  & \small \multirow{-2}{*}{\textbf{\textcolor[HTML]{ff9e00}{1.2}}} \\
    \hline  
  \end{tabular}
  \caption{MuST-SHE gender scores}
  \label{tab:mustshe}
 \end{adjustwidth}
\end{table*}

\newpage
\section{Prompts employed in the Benchmarking} \label{sec:prompts_bench}
The prompts employed with Ǎguila-7B when testing FLoRes-200 devtest set for En $\rightarrow$ Ca and En $\rightarrow$ Es respectively:
\vspace{10pt}
\begin{center}
\begin{adjustwidth}{-0.05cm}{-0.05cm}
\begin{tcolorbox}[colback=white, colframe=black, sharp corners, boxrule=0.5pt, width=1\linewidth]
\setstretch{1.1}
\smaller
\texttt{Translate the following sentence from English to Catalan:} \\
\texttt{English: <s>Hangeul is the only purposely invented alphabet in popular daily use. The alphabet was invented in 1444 during the reign of King Sejong (1418-1450).</s>} \\
\texttt{Catalan: <s>El hangul és l'únic alfabet creat arbitràriament que té un ús estès en la vida diària. L'alfabet es va inventar l'any 1444 durant el regnat de King Sejong (1418-1450).</s>} \\
\texttt{English: <s>They also said in a statement, "The crew is currently working to determine the best method of safely extracting the ship".</s>} \\
\texttt{Catalan: <s>També han dit en un comunicat, "La tripulació treballa ara mateix per a determinar la millor tècnica per a extreure la nau de manera segura".</s>} \\
\texttt{English: <s>This is becoming less of an issue as lens manufacturers achieve higher standards in lens production.</s>} \\
\texttt{Catalan: <s>Això és cada vegada menys important perquè els fabricants de lents estan assolint estàndards més elevats en la producció de lents.</s>} \\
\texttt{English: <s>While assessing the successes and becoming aware of failures, individuals and the whole of the participating persons discover more deeply the values, mission, and driving forces of the organization.</s>} \\
\texttt{Catalan: <s>Mentre confirmen els èxits i prenen consciència dels fracassos, els individus i el grup de participants descobreixen més profundament els valors, la missió i les forces motrius de l'organització.</s>} \\
\texttt{English: <s>Entering Southern Africa by car is an amazing way to see all the region's beauty as well as to get to places off the normal tourist routes.</s>} \\
\texttt{Catalan: <s>Entrar a l'Àfrica del Sud en cotxe és una forma impressionant de veure tota la bellesa de la regió i d'arribar a llocs fora de les rutes turístiques més habituals.</s>} \\
\texttt{English: <s>\_\_\_\_sentence\_to\_translate\_\_\_\_</s>} \\
\texttt{Catalan: <s>}
\end{tcolorbox}
\end{adjustwidth}
\end{center}

\newpage
\vspace{5pt}
\begin{center}
\begin{adjustwidth}{-0.05cm}{-0.05cm}
\begin{tcolorbox}[colback=white, colframe=black, sharp corners, boxrule=0.5pt, width=1\linewidth]
\setstretch{1.1}
\smaller
\texttt{Translate the following sentence from English to Spanish:} \\ 
\texttt{English: <s>Hangeul is the only purposely invented alphabet in popular daily use. The alphabet was invented in 1444 during the reign of King Sejong (1418-1450).</s>} \\
\texttt{Spanish: <s>El alfabeto coreano es el único diseñado en forma deliberada que aún se utiliza a diario popularmente. Se inventó en 1444, durante el reinado de Sejong (1418 a 1450).</s>} \\
\texttt{English: <s>They also said in a statement, "The crew is currently working to determine the best method of safely extracting the ship".</s>} \\
\texttt{Spanish: <s>También se dijo en un comunicado que: «La tripulación se encuentra actualmente trabajando para decidir cuál es el método más seguro para extraer el barco».</s>} \\
\texttt{English: <s>This is becoming less of an issue as lens manufacturers achieve higher standards in lens production.</s>} \\
\texttt{Spanish: <s>Este problema cada vez es menos importante gracias a que los fabricantes de lentes logran estándares más altos en su producción.</s>} \\
\texttt{English: <s>While assessing the successes and becoming aware of failures, individuals and the whole of the participating persons discover more deeply the values, mission, and driving forces of the organization.</s>} \\
\texttt{Spanish: <s>Durante el proceso de análisis de los éxitos y toma de conciencia de los fracasos, los individuos y grupos de personas involucrados descubren con mayor profundidad los valores, el objetivo y las fuerzas que impulsan a la organización.</s>} \\
\texttt{English: <s>Entering Southern Africa by car is an amazing way to see all the region's beauty as well as to get to places off the normal tourist routes.</s>} \\
\texttt{Spanish: <s>Una fantástica forma de contemplar todo el encanto de la región del sur África es ingresar en automóvil, lo que, a su vez, le permitirá acceder a lugares fuera de las rutas turísticas habituales.</s>} \\
\texttt{English: <s>\_\_\_\_sentence\_to\_translate\_\_\_\_</s>} \\
\texttt{Spanish: <s> }
\end{tcolorbox}
\end{adjustwidth}
\end{center}

\newpage
The prompts employed with Ǎguila-7B when testing FLoRes-200 dev set, WinoMT, Gold BUG and MuST-SHE for En $\rightarrow$ Ca and En $\rightarrow$ Es were:
\vspace{10pt}
\begin{center}
\begin{adjustwidth}{-0.05cm}{-0.05cm}
\begin{tcolorbox}[colback=white, colframe=black, sharp corners, boxrule=0.5pt, width=1\linewidth]
\setstretch{1.1}
\smaller
\texttt{Translate the following sentence from English to Catalan:} \\
\texttt{English: <s>The feathers' structure suggests that they were not used in flight but rather for temperature regulation or display. The researchers suggested that, even though this is the tail of a young dinosaur, the sample shows adult plumage and not a chick's down.</s>} \\
\texttt{Catalan: <s>L'estructura de les plomes fa pensar que no s'usaven per a volar sinó per a regular la temperatura o per a exhibir-se. Els investigadors han suggerit que, tot i que es tracta de la cua d'un dinosaure jove, la mostra presenta el plomatge d'un adult i no d'un pollet.</s>} \\
\texttt{English: <s>They found the Sun operated on the same basic principles as other stars: The activity of all stars in the system was found to be driven by their luminosity, their rotation, and nothing else.</s>} \\
\texttt{Catalan: <s>Han descobert que el Sol funcionava sota els mateixos principis bàsics que altres estrelles: s'ha vist que l'activitat de totes les estrelles del sistema depèn de llur brillantor, llur rotació i res més.</s>} \\
\texttt{English: <s>The speeds of 802.11n are substantially faster than that of its predecessors with a maximum theoretical throughput of 600Mbit/s.</s>} \\
\texttt{Catalan: <s>Les velocitats de 802.11n són substancialment més ràpides que les dels seus predecessors amb un rendiment teòric màxim de 600Mbit/s.</s>} \\
\texttt{English: <s>Over four million people went to Rome to attend the funeral.</s>} \\
\texttt{Catalan: <s>Més de quatre milions de persones van anar a Roma per a assistir al funeral.</s>} \\
\texttt{English: <s>Mrs. Kirchner announced her intention to run for president at the Argentine Theatre, the same location she used to start her 2005 campaign for the Senate as member of the Buenos Aires province delegation.</s>} \\
\texttt{Catalan: <s>La Sra. Kirchner va anunciar la seva intenció de presentar-se a la presidència al Teatre de l'Argentina, el mateix lloc on va engegar la campanya al Senat de 2005 com a membre de la delegació provincial de Buenos Aires.</s>} \\
\texttt{English: <s>\_\_\_\_sentence\_to\_translate\_\_\_\_</s>} \\
\texttt{Catalan: <s>}
\end{tcolorbox}
\end{adjustwidth}
\end{center}

\newpage
\vspace{5pt}
\begin{center}
\begin{adjustwidth}{-0.05cm}{-0.05cm}
\begin{tcolorbox}[colback=white, colframe=black, sharp corners, boxrule=0.5pt, width=1\linewidth]
\setstretch{1.1}
\smaller
\texttt{Translate the following sentence from English to Spanish:} \\
\texttt{English: <s>The feathers' structure suggests that they were not used in flight but rather for temperature regulation or display. The researchers suggested that, even though this is the tail of a young dinosaur, the sample shows adult plumage and not a chick's down.</s>} \\
\texttt{Spanish: <s>La estructura que presenta el plumaje sugiere que su función no estaba relacionada con el vuelo, sino que las usaban para regular la temperatura o como indicador de la misma. Los investigadores sostienen que, aunque se trata de la cola de un dinosaurio joven, la muestra analizada presenta rasgos del plumaje de un adulto y no de un polluelo.</s>} \\
\texttt{English: <s>They found the Sun operated on the same basic principles as other stars: The activity of all stars in the system was found to be driven by their luminosity, their rotation, and nothing else.</s>} \\
\texttt{Spanish: <s>Se descubrió que el sol se regía por los mismos principios básicos que otras estrellas: los únicos factores que impulsaban su actividad dentro del sistema eran su luminosidad y su rotación.</s>} \\
\texttt{English: <s>The speeds of 802.11n are substantially faster than that of its predecessors with a maximum theoretical throughput of 600Mbit/s.</s>} \\
\texttt{Spanish: <s>Las velocidades del estándar 802.11n son mucho más altas que las alcanzadas por los que lo precedieron, con un rendimiento teórico máximo de 600 Mbps.</s>} \\
\texttt{English: <s>Over four million people went to Rome to attend the funeral.</s>} \\
\texttt{Spanish: <s>Más de cuatro millones de individuos se concentraron en Roma para presenciar el funeral.</s>} \\
\texttt{English: <s>Mrs. Kirchner announced her intention to run for president at the Argentine Theatre, the same location she used to start her 2005 campaign for the Senate as member of the Buenos Aires province delegation.</s>} \\
\texttt{Spanish: <s>El Teatro Argentino fue el lugar donde la señora Kirchner anunció su intención de candidatearse como presidenta; este es el mismo sitio donde inició su campaña para el senado en el año 2005, en representación de la provincia de Buenos Aires.</s>} \\
\texttt{English: <s>\_\_\_\_sentence\_to\_translate\_\_\_\_</s>} \\
\texttt{Spanish: <s>}
\end{tcolorbox}
\end{adjustwidth}
\end{center}

\newpage
The prompts employed with Flor-6.3B and Llama-2-7B when testing FLoRes-200 devtest set for En $\rightarrow$ Ca and En $\rightarrow$ Es respectively:
\vspace{10pt}
\begin{center}
\begin{adjustwidth}{-0.05cm}{-0.05cm}
\begin{tcolorbox}[colback=white, colframe=black, sharp corners, boxrule=0.5pt, width=1\linewidth]
\setstretch{1.1}
\smaller
\texttt{Translate the following sentence from English to Catalan:} \\
\texttt{English: <BOS>Hangeul is the only purposely invented alphabet in popular daily use. The alphabet was invented in 1444 during the reign of King Sejong (1418-1450).<EOS>} \\
\texttt{Catalan: <BOS>El hangul és l'únic alfabet creat arbitràriament que té un ús estès en la vida diària. L'alfabet es va inventar l'any 1444 durant el regnat de King Sejong (1418-1450).<EOS>} \\
\texttt{English: <BOS>They also said in a statement, "The crew is currently working to determine the best method of safely extracting the ship".<EOS>} \\
\texttt{Catalan: <BOS>També han dit en un comunicat, "La tripulació treballa ara mateix per a determinar la millor tècnica per a extreure la nau de manera segura".<EOS>} \\
\texttt{English: <BOS>This is becoming less of an issue as lens manufacturers achieve higher standards in lens production.<EOS>} \\
\texttt{Catalan: <BOS>Això és cada vegada menys important perquè els fabricants de lents estan assolint estàndards més elevats en la producció de lents.<EOS>} \\
\texttt{English: <BOS>While assessing the successes and becoming aware of failures, individuals and the whole of the participating persons discover more deeply the values, mission, and driving forces of the organization.<EOS>} \\
\texttt{Catalan: <BOS>Mentre confirmen els èxits i prenen consciència dels fracassos, els individus i el grup de participants descobreixen més profundament els valors, la missió i les forces motrius de l'organització.<EOS>} \\
\texttt{English: <BOS>Entering Southern Africa by car is an amazing way to see all the region's beauty as well as to get to places off the normal tourist routes.<EOS>} \\
\texttt{Catalan: <BOS>Entrar a l'Àfrica del Sud en cotxe és una forma impressionant de veure tota la bellesa de la regió i d'arribar a llocs fora de les rutes turístiques més habituals.<EOS>} \\
\texttt{English: <BOS>\_\_\_\_sentence\_to\_translate\_\_\_\_<EOS>} \\
\texttt{Catalan: <BOS>}  
\end{tcolorbox}
\end{adjustwidth}
\end{center}

\newpage
\vspace{5pt}
\begin{center}
\begin{adjustwidth}{-0.05cm}{-0.05cm}
\begin{tcolorbox}[colback=white, colframe=black, sharp corners, boxrule=0.5pt, width=1\linewidth]
\setstretch{1.1}
\smaller
\texttt{Translate the following sentence from English to Spanish:} \\
\texttt{English: <BOS>Hangeul is the only purposely invented alphabet in popular daily use. The alphabet was invented in 1444 during the reign of King Sejong (1418-1450).<EOS>} \\
\texttt{Spanish: <BOS>El alfabeto coreano es el único diseñado en forma deliberada que aún se utiliza a diario popularmente. Se inventó en 1444, durante el reinado de Sejong (1418 a 1450).<EOS>} \\
\texttt{English: <BOS>They also said in a statement, "The crew is currently working to determine the best method of safely extracting the ship".<EOS>} \\
\texttt{Spanish: <BOS>También se dijo en un comunicado que: «La tripulación se encuentra actualmente trabajando para decidir cuál es el método más seguro para extraer el barco».<EOS>} \\
\texttt{English: <BOS>This is becoming less of an issue as lens manufacturers achieve higher standards in lens production.<EOS>} \\
\texttt{Spanish: <BOS>Este problema cada vez es menos importante gracias a que los fabricantes de lentes logran estándares más altos en su producción.<EOS>} \\
\texttt{English: <BOS>While assessing the successes and becoming aware of failures, individuals and the whole of the participating persons discover more deeply the values, mission, and driving forces of the organization.<EOS>} \\
\texttt{Spanish: <BOS>Durante el proceso de análisis de los éxitos y toma de conciencia de los fracasos, los individuos y grupos de personas involucrados descubren con mayor profundidad los valores, el objetivo y las fuerzas que impulsan a la organización.<EOS>} \\
\texttt{English: <BOS>Entering Southern Africa by car is an amazing way to see all the region's beauty as well as to get to places off the normal tourist routes.<EOS>} \\
\texttt{Spanish: <EOS>Una fantástica forma de contemplar todo el encanto de la región del sur África es ingresar en automóvil, lo que, a su vez, le permitirá acceder a lugares fuera de las rutas turísticas habituales.<BOS>} \\
\texttt{English: <BOS>\_\_\_\_sentence\_to\_translate\_\_\_\_<EOS>} \\
\texttt{Spanish: <BOS>}
\end{tcolorbox}
\end{adjustwidth}
\end{center}

\newpage
The prompts employed with Flor-6.3B and Llama-2-7B when testing FLoRes-200 dev set, WinoMT, Gold BUG and MuST-SHE for En $\rightarrow$ Ca and En $\rightarrow$ Es were: 
\vspace{5pt}
\begin{center}
\begin{adjustwidth}{-0.05cm}{-0.05cm}
\begin{tcolorbox}[colback=white, colframe=black, sharp corners, boxrule=0.5pt, width=1\linewidth]
\setstretch{1.1}
\smaller
\texttt{Translate the following sentence from English to Catalan:} \\
\texttt{English: <BOS>The feathers' structure suggests that they were not used in flight but rather for temperature regulation or display. The researchers suggested that, even though this is the tail of a young dinosaur, the sample shows adult plumage and not a chick's down.<EOS>} \\
\texttt{Catalan: <BOS>L'estructura de les plomes fa pensar que no s'usaven per a volar sinó per a regular la temperatura o per a exhibir-se. Els investigadors han suggerit que, tot i que es tracta de la cua d'un dinosaure jove, la mostra presenta el plomatge d'un adult i no d'un pollet.<EOS>} \\
\texttt{English: <BOS>They found the Sun operated on the same basic principles as other stars: The activity of all stars in the system was found to be driven by their luminosity, their rotation, and nothing else.<EOS>} \\
\texttt{Catalan: <BOS>Han descobert que el Sol funcionava sota els mateixos principis bàsics que altres estrelles: s'ha vist que l'activitat de totes les estrelles del sistema depèn de llur brillantor, llur rotació i res més.<EOS>} \\
\texttt{English: <BOS>The speeds of 802.11n are substantially faster than that of its predecessors with a maximum theoretical throughput of 600Mbit/s.<EOS>} \\
\texttt{Catalan: <BOS>Les velocitats de 802.11n són substancialment més ràpides que les dels seus predecessors amb un rendiment teòric màxim de 600Mbit/s.<EOS>} \\
\texttt{English: <BOS>Over four million people went to Rome to attend the funeral.<EOS>} \\
\texttt{Catalan: <BOS>Més de quatre milions de persones van anar a Roma per a assistir al funeral.<EOS>} \\
\texttt{English: <BOS>Mrs. Kirchner announced her intention to run for president at the Argentine Theatre, the same location she used to start her 2005 campaign for the Senate as member of the Buenos Aires province delegation.<EOS>} \\
\texttt{Catalan: <BOS>La Sra. Kirchner va anunciar la seva intenció de presentar-se a la presidència al Teatre de l'Argentina, el mateix lloc on va engegar la campanya al Senat de 2005 com a membre de la delegació provincial de Buenos Aires.<EOS>} \\
\texttt{English: <BOS>\_\_\_\_sentence\_to\_translate\_\_\_\_<EOS>} \\
\texttt{Catalan: <BOS>}
\end{tcolorbox}
\end{adjustwidth}
\end{center}

\newpage
\vspace{5pt}
\begin{center}
\begin{adjustwidth}{-0.05cm}{-0.05cm}
\begin{tcolorbox}[colback=white, colframe=black, sharp corners, boxrule=0.5pt, width=1\linewidth]
\setstretch{1.1}
\smaller
\texttt{Translate the following sentence from English to Spanish:} \\
\texttt{English: <BOS>The feathers' structure suggests that they were not used in flight but rather for temperature regulation or display. The researchers suggested that, even though this is the tail of a young dinosaur, the sample shows adult plumage and not a chick's down.<EOS>} \\
\texttt{Spanish: <BOS>La estructura que presenta el plumaje sugiere que su función no estaba relacionada con el vuelo, sino que las usaban para regular la temperatura o como indicador de la misma. Los investigadores sostienen que, aunque se trata de la cola de un dinosaurio joven, la muestra analizada presenta rasgos del plumaje de un adulto y no de un polluelo.<EOS>} \\
\texttt{English: <BOS>They found the Sun operated on the same basic principles as other stars: The activity of all stars in the system was found to be driven by their luminosity, their rotation, and nothing else.<EOS>} \\
\texttt{Spanish: <BOS>Se descubrió que el sol se regía por los mismos principios básicos que otras estrellas: los únicos factores que impulsaban su actividad dentro del sistema eran su luminosidad y su rotación.<EOS>} \\
\texttt{English: <BOS>The speeds of 802.11n are substantially faster than that of its predecessors with a maximum theoretical throughput of 600Mbit/s.<EOS>} \\
\texttt{Spanish: <BOS>Las velocidades del estándar 802.11n son mucho más altas que las alcanzadas por los que lo precedieron, con un rendimiento teórico máximo de 600 Mbps.<EOS>} \\
\texttt{English: <BOS>Over four million people went to Rome to attend the funeral.<EOS>} \\
\texttt{Spanish: <BOS>Más de cuatro millones de individuos se concentraron en Roma para presenciar el funeral.<EOS>} \\
\texttt{English: <BOS>Mrs. Kirchner announced her intention to run for president at the Argentine Theatre, the same location she used to start her 2005 campaign for the Senate as member of the Buenos Aires province delegation.<EOS>} \\
\texttt{Spanish: <BOS>El Teatro Argentino fue el lugar donde la señora Kirchner anunció su intención de candidatearse como presidenta; este es el mismo sitio donde inició su campaña para el senado en el año 2005, en representación de la provincia de Buenos Aires.<EOS>} \\
\texttt{English: <BOS>\_\_\_\_sentence\_to\_translate\_\_\_\_<EOS>} \\
\texttt{Spanish: <BOS>}
\end{tcolorbox}
\end{adjustwidth}
\end{center}

\newpage
\section{Gender Scores on the Pro- and Anti-Stereotypical sets from WinoMT and Gold BUG} \label{sec:pro_and_anti}
Below you can see the results for the WinoMT: \\
\begin{table*}[h]
 \begin{adjustwidth}{-1cm}{-1cm}
  \centering
  \renewcommand{\arraystretch}{1.2} 
  \begin{tabular}{|c|c|c|c|c|c|c|c|c|c|}
    \cline{3-10}
    \multicolumn{2}{c|}{} & \multicolumn{4}{|c|}{English $\rightarrow$ Catalan} & \multicolumn{4}{|c|}{English $\rightarrow$ Spanish} \\
    \cline{3-10}
    \multicolumn{2}{c|}{} & \small \textbf{G Acc} & \small \textbf{F1-male} & \small \textbf{F1-female} & \small \textbf{$\Delta$G} & \small \textbf{G Acc} & \small \textbf{F1-male} & \small \textbf{F1-female} & \small \textbf{$\Delta$G} \\
    \hline
    \multirow{4}{*}{\rotatebox[origin=c]{90}{NMT}} & \cellcolor[HTML]{C9DAF8} \small Google Translate &   \small 74.1 & \small 80.9 & \small 71.1 & \small 9.8 & \small \textbf{\textcolor{blue}{89.8}} & \small \textbf{\textcolor{blue}{91.1}} & \small \textbf{\textcolor{blue}{90.4}} & \small \textbf{\textcolor{blue}{0.7}} \\
    \cline{2-10}
     & \cellcolor[HTML]{C9DAF8} \small NLLB-200-1.3B & \small \textbf{\textcolor{blue}{79.7}} & \small \textbf{\textcolor{blue}{85.1}} &  \small \textbf{\textcolor{blue}{80.7}} &  \small \textbf{\textcolor{blue}{4.4}} & \small 89.3 & \small 90.8 & \small 89.8 & \small 1.0 \\
    \cline{2-10}
     & \cellcolor[HTML]{C9DAF8} \small M2M-100-1.2B & \small 67.7 & \small 76.4 & \small 61.1 & \small 15.3 & \small 73.7 & \small 79.2 & \small 68.7 & \small 10.5 \\
    \cline{2-10}
     & \cellcolor[HTML]{C9DAF8} \small Mt-aina-en-ca & \small 65.7 & \small 76.0 & \small 56.9 & \small 19.1  & - & - & - & -  \\
    \hline
    \hline
    \multirow{3}{*}{\rotatebox[origin=c]{90}{LLM}}
     &  \cellcolor[HTML]{F4CCCC}  \small  Ǎguila-7B &  \small 66.0 & \small 76.1 & \small  \textbf{\textcolor{red}{57.8}} & \small  \textbf{\textcolor{red}{18.3}} & \small 65.7 & \small 74.1 & \small 53.8 & \small 20.3  \\
    \cline{2-10}
    &  \cellcolor[HTML]{F4CCCC}  \small  Flor-6.3B & \small 66.4 & \small 76.3 & \small 57.6 & \small 18.7 & \small 71.9 & \small 77.9 & \small 63.3 & \small 14.6 \\
    \cline{2-10}
     &  \cellcolor[HTML]{F4CCCC}  \small Llama-2-7B &  \small \textbf{\textcolor{red}{66.5}} &   \small \textbf{\textcolor{red}{78.0}} &  \small 57.7 &   \small 20.3 & \small \textbf{\textcolor{red}{73.1}} &   \small \textbf{\textcolor{red}{78.5}} &  \small \textbf{\textcolor{red}{66.4}} & \small \textbf{\textcolor{red}{12.1}}\\
    \hline  
  \end{tabular}
  \caption{WinoMT pro-stereotypical set gender scores}
  \label{tab:winomt_pro}
 \end{adjustwidth}
\end{table*}
\begin{table*}[h]
 \begin{adjustwidth}{-1cm}{-1cm}
  \centering
  \renewcommand{\arraystretch}{1.2} 
  \begin{tabular}{|c|c|c|c|c|c|c|c|c|c|}
    \cline{3-10}
    \multicolumn{2}{c|}{} & \multicolumn{4}{|c|}{English $\rightarrow$ Catalan} & \multicolumn{4}{|c|}{English $\rightarrow$ Spanish} \\
    \cline{3-10}
    \multicolumn{2}{c|}{} & \small \textbf{G Acc} & \small \textbf{F1-male} & \small \textbf{F1-female} & \small \textbf{$\Delta$G} & \small \textbf{G Acc} & \small \textbf{F1-male} & \small \textbf{F1-female} & \small \textbf{$\Delta$G} \\
    \hline
    \multirow{4}{*}{\rotatebox[origin=c]{90}{NMT}}  & \small \cellcolor[HTML]{C9DAF8} Google Translate &   \small 51.8 & \small 60.6 & \small 43.7 & \small 16.9 & \small \textbf{\textcolor{blue}{66.9}} & \small \textbf{\textcolor{blue}{72.4}} & \small \textbf{\textcolor{blue}{60.6}} & \small \textbf{\textcolor{blue}{11.8}} \\
    \cline{2-10}
     &  \small \cellcolor[HTML]{C9DAF8} NLLB-200-1.3B & \small \textbf{\textcolor{blue}{53.0}} &   \small \textbf{\textcolor{blue}{62.3}} &  \small \textbf{\textcolor{blue}{47.3}} &   \small \textbf{\textcolor{blue}{15.0}} & \small 58.1 & \small 66.7 & \small 46.2 & \small 20.5 \\
    \cline{2-10}
     &  \small \cellcolor[HTML]{C9DAF8} M2M-100-1.2B & \small 45.9 & \small 58.3 & \small 30.0 & \small 28.3 & \small 52.5 & \small 65.3 & \small 29.6 & \small 35.7 \\
    \cline{2-10}
     &  \small \cellcolor[HTML]{C9DAF8} Mt-aina-en-ca & \small 42.5 & \small 56.8 & \small 21.5 & \small 35.3 & - & - & - & -  \\
    \hline
    \hline
    \multirow{3}{*}{\rotatebox[origin=c]{90}{LLM}}
     & \small  \cellcolor[HTML]{F4CCCC}  Ǎguila-7B &  \small 34.5  &  \small 49.8 & \small 12.0 & \small 37.8 & \small 43.1 & \small 58.5 & \small 12.7 & \small 45.8 \\
    \cline{2-10}
    & \small  \cellcolor[HTML]{F4CCCC}  Flor-6.3B &  \small 38.7 & \small \textbf{\textcolor{red}{54.0}} & \small 13.8 & \small 40.2 & \small 45.2  & \small 58.2 & \small \textbf{\textcolor{red}{22.8}} & \small \textbf{\textcolor{red}{35.4}} \\
    \cline{2-10}
     &  \small  \cellcolor[HTML]{F4CCCC} Llama-2-7B &  \small \textbf{\textcolor{red}{38.8}} &   \small 53.6 &  \small \textbf{\textcolor{red}{16.6}} &   \small \textbf{\textcolor{red}{37.0}} & \small \textbf{\textcolor{red}{45.3}} &   \small \textbf{\textcolor{red}{59.0}} &  \small 19.7 & \small 39.3 \\
    \hline  
  \end{tabular}
  \caption{WinoMT anti-stereotypical set gender scores}
  \label{tab:winomy_anti}
 \end{adjustwidth}
\end{table*}

\newpage
Below you can see the results for the Gold BUG: \\
\begin{table*}[h]
 \begin{adjustwidth}{-1cm}{-1cm}
  \centering
  \renewcommand{\arraystretch}{1.2} 
  \begin{tabular}{|c|c|c|c|c|c|c|c|c|c|}
    \cline{3-10}
    \multicolumn{2}{c|}{} & \multicolumn{4}{|c|}{English $\rightarrow$ Catalan} & \multicolumn{4}{|c|}{English $\rightarrow$ Spanish} \\
    \cline{3-10}
    \multicolumn{2}{c|}{} & \small \textbf{G Acc} & \small \textbf{F1-male} & \small \textbf{F1-female} & \small \textbf{$\Delta$G} & \small \textbf{G Acc} & \small \textbf{F1-male} & \small \textbf{F1-female} & \small \textbf{$\Delta$G} \\
    \hline
    \multirow{4}{*}{\rotatebox[origin=c]{90}{NMT}} &  \cellcolor[HTML]{C9DAF8} \small Google Translate &  \small \textbf{\textcolor{blue}{69.6}} &  \small \textbf{\textcolor{blue}{82.6}} &  \small \textbf{\textcolor{blue}{67.5}} & \small \textbf{\textcolor{blue}{15.1}} & \small 70.8 & \small 83.9 & \small 60.5 & \small 23.4 \\
    \cline{2-10}
     &   \cellcolor[HTML]{C9DAF8} \small NLLB-200-1.3B & \small 66.9 & \small 81.3 & \small 52.7 &   \small 28.6 & \small \textbf{\textcolor{blue}{71.5}} & \small \textbf{\textcolor{blue}{84.1}} & \small \textbf{\textcolor{blue}{66.6}} & \small \textbf{\textcolor{blue}{17.5}} \\
    \cline{2-10}
     &  \cellcolor[HTML]{C9DAF8} \small M2M-100-1.2B & \small 67.4 & \small 81.8 & \small 54.7 & \small 27.1 & \small 70.3 & \small 83.6 & \small 61.3 & \small 22.3 \\
    \cline{2-10}
     &  \cellcolor[HTML]{C9DAF8} \small Mt-aina-en-ca & \small 68.0 & \small 81.8 & \small 64.4 & \small  17.4 & - & - & - & -  \\
    \hline
    \hline
    \multirow{3}{*}{\rotatebox[origin=c]{90}{LLM}}
     & \cellcolor[HTML]{F4CCCC} \small  Ǎguila-7B & \small 60.7 & \small 76.3 & \small 54.8 & \small \textbf{\textcolor{red}{21.5}} & \small 64.4 & \small 79.1 & \small \textbf{\textcolor{red}{56.0}} & \small \textbf{\textcolor{red}{23.1}} \\
    \cline{2-10}
    & \cellcolor[HTML]{F4CCCC} \small  Flor-6.3B &  \small  65.0 & \small 80.1 & \small 54.7 & \small 25.4 & \small 69.8  & \small 83.2 & \small 54.6 & \small 28.6 \\
    \cline{2-10}
     & \cellcolor[HTML]{F4CCCC} \small Llama-2-7B &  \small \textbf{\textcolor{red}{66.5}} &   \small \textbf{\textcolor{red}{81.3}} &  \small \textbf{\textcolor{red}{56.1}} &   \small 25.2 & \small \textbf{\textcolor{red}{69.9}} & \small \textbf{\textcolor{red}{83.4}} & \small 53.1 & \small 30.3 \\
    \hline  
  \end{tabular}
  \caption{Gold BUG pro-stereotypical set gender scores}
  \label{tab:goldbug_pro}
 \end{adjustwidth}
\end{table*}

\begin{table*}[h]
 \begin{adjustwidth}{-1cm}{-1cm}
  \centering
  \renewcommand{\arraystretch}{1.2} 
  \begin{tabular}{|c|c|c|c|c|c|c|c|c|c|}
    \cline{3-10}
    \multicolumn{2}{c|}{} & \multicolumn{4}{|c|}{English $\rightarrow$ Catalan} & \multicolumn{4}{|c|}{English $\rightarrow$ Spanish} \\
    \cline{3-10}
    \multicolumn{2}{c|}{} & \small \textbf{G Acc} & \small \textbf{F1-male} & \small \textbf{F1-female} & \small \textbf{$\Delta$G} & \small \textbf{G Acc} & \small \textbf{F1-male} & \small \textbf{F1-female} & \small \textbf{$\Delta$G} \\
    \hline
    \multirow{4}{*}{\rotatebox[origin=c]{90}{NMT}} &  \cellcolor[HTML]{C9DAF8} \small Google Translate &   \small 43.6 &  \small 61.0 &  \small 35.7 & \small 25.3 & \small \textbf{\textcolor{blue}{51.4}} & \small \textbf{\textcolor{blue}{67.1}} & \small \textbf{\textcolor{blue}{47.6}} & \small \textbf{\textcolor{blue}{19.5}} \\
    \cline{2-10}
     &  \cellcolor[HTML]{C9DAF8} \small NLLB-200-1.3B & \small \textbf{\textcolor{blue}{46.9}} & \small \textbf{\textcolor{blue}{62.9}} &  \small \textbf{\textcolor{blue}{44.0}} &   \small 18.9 & \small 48.8 & \small 65.5 & \small 44.4 & \small 21.1 \\
    \cline{2-10}
     &  \cellcolor[HTML]{C9DAF8} \small M2M-100-1.2B & \small 41.9 & \small 59.5 & \small 29.2 & \small 30.3 & \small 46.2 & \small 62.8 & \small 36.8 & \small 26.0 \\
    \cline{2-10}
     &  \cellcolor[HTML]{C9DAF8} \small Mt-aina-en-ca & \small 46.0 & \small 61.3 & \small 43.9 & \small \textbf{\textcolor{blue}{17.4}} & - & - & - & -  \\
    \hline
    \hline
    \multirow{3}{*}{\rotatebox[origin=c]{90}{LLM}}
     & \cellcolor[HTML]{F4CCCC} \small  Ǎguila-7B & \small 40.0 & \small 59.1 & \small 27.0 & \small 32.1 & \small 46.0 & \small 64.8 &  \small 32.8 & \small 32.0 \\
    \cline{2-10}
    & \cellcolor[HTML]{F4CCCC} \small  Flor-6.3B &  \small 44.5 & \small 62.5 & \small 35.1 & \small \textbf{\textcolor{red}{27.4}} & \small 49.0 & \small 66.2 & \small \textbf{\textcolor{red}{41.9}} & \small \textbf{\textcolor{red}{24.3}} \\
    \cline{2-10}
     &  \cellcolor[HTML]{F4CCCC} \small Llama-2-7B &  \small \textbf{\textcolor{red}{46.7}} & \small \textbf{\textcolor{red}{64.4}} & \small \textbf{\textcolor{red}{35.7}} & \small 28.7 & \small \textbf{\textcolor{red}{49.8}} & \small \textbf{\textcolor{red}{69.0}} &  \small 35.4 & \small 33.6 \\
    \hline  
  \end{tabular}
  \caption{Gold BUG anti-stereotypical set gender scores}
  \label{tab:goldbug_anti}
 \end{adjustwidth}
\end{table*}

\newpage
\section{Proportion of Predicted Male and Female terms in Absence of Gender Cues} \label{sec:no_context}
The following Figures \ref{fig:pie_charts_ca} and \ref{fig:pie_charts_es} depict a range of pie diagrams illustrating the proportion of predicted male and female terms in the translations per model when testing on instances of MuST-SHE without gender cues for disambiguation. \\

\begin{figure}[H]
    \centering
    \makebox[\textwidth][c]{\includegraphics[width=1\textwidth]{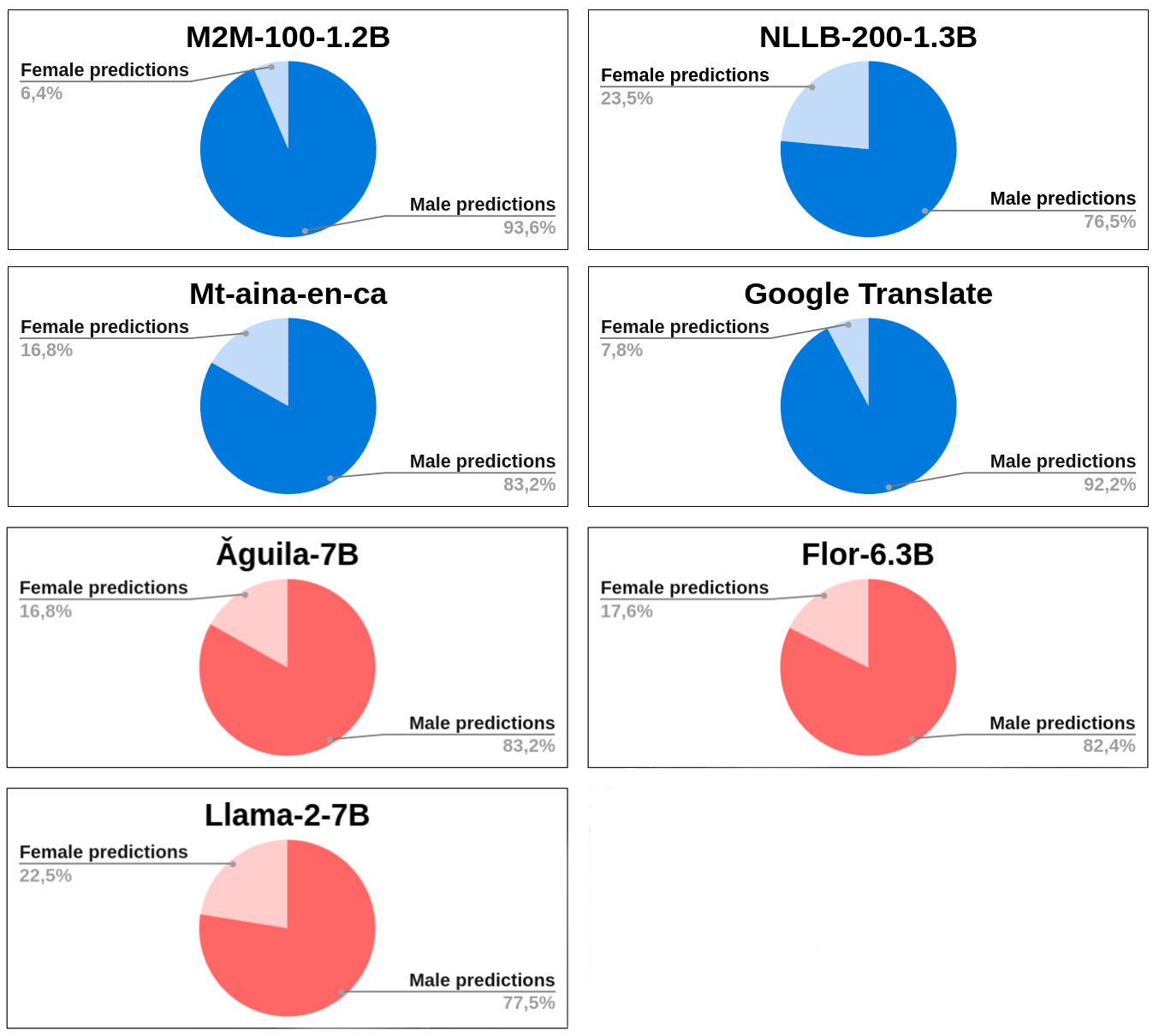}}
    %\captionsetup{width=0.7\textwidth}
    \caption{Male and female predicted terms across models for En $\rightarrow$ Ca in absence of gender cues}
    \label{fig:pie_charts_ca}
\end{figure}

\newpage

\begin{figure}[H]
    \centering
    \makebox[\textwidth][c]{\includegraphics[width=1\textwidth]{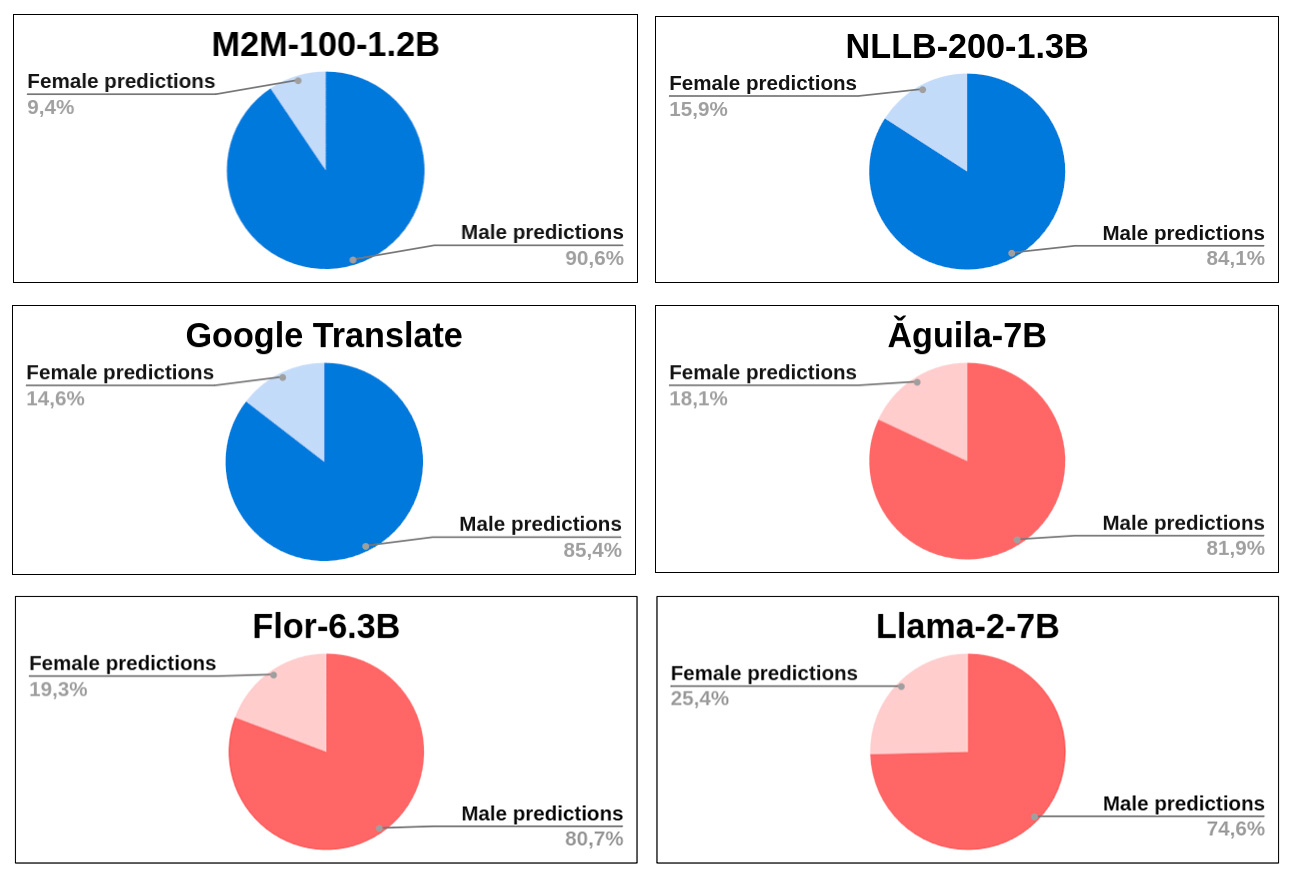}}
    \caption{Male and female predicted terms across models for En $\rightarrow$ Es in absence of gender cues}
    \label{fig:pie_charts_es}
\end{figure}

\newpage
\section{Prompt used for the Baseline in the Investigation into Prompting} \label{sec:prompts_llama2chat}

An example of the resulting prompt used for Llama-2-7B-chat after the format adaptations: 
\vspace{5pt}
\begin{center}
\begin{adjustwidth}{-0.05cm}{-0.05cm}
\begin{tcolorbox}[colback=white, colframe=black, sharp corners, boxrule=0.5pt, width=1\linewidth]
\setstretch{1.1}
\smaller
\texttt{<<SYS>> Translate the following sentence from English to Catalan: <</SYS>>} \\
\texttt{[INST] English: <BOS>The feathers' structure suggests that they were not used in flight but rather for temperature regulation or display. The researchers suggested that, even though this is the tail of a young dinosaur, the sample shows adult plumage and not a chick's down.<EOS> [/INST]} \\
\texttt{Catalan: <BOS>L'estructura de les plomes fa pensar que no s'usaven per a volar sinó per a regular la temperatura o per a exhibir-se. Els investigadors han suggerit que, tot i que es tracta de la cua d'un dinosaure jove, la mostra presenta el plomatge d'un adult i no d'un pollet.<EOS>} \\
\texttt{[INST] English: <BOS>They found the Sun operated on the same basic principles as other stars: The activity of all stars in the system was found to be driven by their luminosity, their rotation, and nothing else.<EOS> [/INST]} \\
\texttt{Catalan: <BOS>Han descobert que el Sol funcionava sota els mateixos principis bàsics que altres estrelles: s'ha vist que l'activitat de totes les estrelles del sistema depèn de llur brillantor, llur rotació i res més.<EOS>} \\
\texttt{[INST] English: <BOS>The speeds of 802.11n are substantially faster than that of its predecessors with a maximum theoretical throughput of 600Mbit/s.<EOS> [/INST]} \\
\texttt{Catalan: <BOS>Les velocitats de 802.11n són substancialment més ràpides que les dels seus predecessors amb un rendiment teòric màxim de 600Mbit/s.<EOS>} \\
\texttt{[INST] English: <BOS>Over four million people went to Rome to attend the funeral.<EOS> [/INST]} \\
\texttt{Catalan: <BOS>Més de quatre milions de persones van anar a Roma per a assistir al funeral.<EOS>} \\
\texttt{[INST] English: <BOS>Mrs. Kirchner announced her intention to run for president at the Argentine Theatre, the same location she used to start her 2005 campaign for the Senate as member of the Buenos Aires province delegation.<EOS> [/INST]} \\
\texttt{Catalan: <BOS>La Sra. Kirchner va anunciar la seva intenció de presentar-se a la presidència al Teatre de l'Argentina, el mateix lloc on va engegar la campanya al Senat de 2005 com a membre de la delegació provincial de Buenos Aires.<EOS>} \\
\texttt{[INST] English: <BOS>\_\_\_\_sentence\_to\_translate\_\_\_\_<EOS> [/INST]} \\
\texttt{Catalan: <BOS>} 
\end{tcolorbox}
\end{adjustwidth}
\end{center}

\newpage
\section{Curated Prompts for the Investigation into Prompting} \label{sec:prompts_inv}
Below are all the different prompts used with Llama-2-7B-chat that have been tested on WinoMT test set. \\
\\
Prompt with 5-shot MuST-SHE examples:
\begin{center}
\begin{adjustwidth}{-0.05cm}{-0.05cm}
\begin{tcolorbox}[colback=white, colframe=black, sharp corners, boxrule=0.5pt, width=1\linewidth]
\setstretch{1.1}
\smaller
\texttt{<<SYS>> Translate the following sentence from English to Catalan: <</SYS>>} \\
\texttt{[INST] English: <BOS>Early on, Laura Hughes could see that I was a little lost in this habitat, so she often sat right next to me in meetings so she could be my tech translator, and I could write her notes and she could tell me, "That's what that means." Laura was 27 years old, she'd worked for Google for four years and then for a year and a half at Airbnb when I met her.<EOS> [/INST]} \\
\texttt{Catalan: <BOS>Al principi, la Laura Hughes va poder veure que estava una mica perdut en aquest hàbitat, així que sovint s'asseia al meu costat a les reunions per poder ser la meva traductora de tecnologia, i jo podia escriure-li notes i ella em podria dir, "Això és el que això significa." La Laura tenia 27 anys, havia treballat a Google durant quatre anys i després durant un any i mig a Airbnb quan la vaig conèixer.<EOS>} \\
\\
\texttt{[INST] English: <BOS>When I found the captain, he was having a very engaging conversation with the homeowner, who was surely having one of the worst days of her life.<EOS> [/INST]} \\
\texttt{Catalan: <BOS>Quan vaig trobar el capità, estava mantenint una conversa molt atractiva amb la propietària, que segurament vivia un dels pitjors dies de la seva vida.<EOS>} \\
\\
\texttt{[INST] English: <BOS>And in this program, girls who have been studying computer skills and the STEM program have a chance to work side by side with young professionals, so that they can learn firsthand what it's like to be an architect, a designer or a scientist.<EOS> [/INST]} \\
\texttt{Catalan: <BOS>I en aquest programa, les noies que han estudiat informàtica i el programa STEM tenen l'oportunitat de treballar colze a colze amb joves professionals, per tal que puguin conèixer de primera mà com és ser una arquitecta, una dissenyadora o una científica.<EOS>} \\
\\
\texttt{[INST] English: <BOS>One government scientist, a friend of mine, we'll call him McPherson, was concerned about the impact government policies were having on his research and the state of science deteriorating in Canada.<EOS> [/INST]} \\
\texttt{Catalan: <BOS>Un científic del govern, un amic meu, l'anomenarem McPherson, estava preocupat per l'impacte que tenien les polítiques governamentals en la seva investigació i el deteriorament de l'estat de la ciència al Canadà.<EOS>} \\
\\
\texttt{[INST] English: <BOS>The architect Emmanuelle Moureaux uses this idea in her work a lot.<EOS> [/INST]} \\
\texttt{Catalan: <BOS>L'arquitecta Emmanuelle Moureaux utilitza molt aquesta idea en la seva obra.<EOS>} \\
\\
\texttt{[INST] English: <BOS>\_\_\_\_sentence\_to\_translate\_\_\_\_<EOS> [/INST]} \\
\texttt{Catalan: <BOS>}
\end{tcolorbox}
\end{adjustwidth}
\end{center}

\newpage
\begin{center}
\begin{adjustwidth}{-0.05cm}{-0.05cm}
\begin{tcolorbox}[colback=white, colframe=black, sharp corners, boxrule=0.5pt, width=1\linewidth]
\setstretch{1.1}
\smaller
\texttt{<<SYS>> Translate the following sentence from English to Spanish: <</SYS>>} \\
\texttt{[INST] English: <BOS>Early on, Laura Hughes could see that I was a little lost in this habitat, so she often sat right next to me in meetings so she could be my tech translator, and I could write her notes and she could tell me, "That's what that means." Laura was 27 years old, she'd worked for Google for four years and then for a year and a half at Airbnb when I met her.<EOS> [/INST]} \\
\texttt{Spanish: <BOS>Al principio, Laura Hughes se dio cuenta de que estaba perdido en este hábitat, así que solía sentarse a mi lado en las reuniones para ser mi traductora de tecnología, y yo le escribía notas y ella me decía, "Esto es lo que significa". Laura tenía 27 años, trabajó en Google durante 4 años, y luego por un año y medio en Airbnb cuando la conocí.<EOS>} \\
\\
\texttt{[INST] English: <BOS>When I found the captain, he was having a very engaging conversation with the homeowner, who was surely having one of the worst days of her life.<EOS> [/INST]} \\
\texttt{Spanish: <BOS>Cuando encontré al capitán, estaba enfrascado en una conversación con la propietaria que sin duda atravesaba uno de los peores días de su vida.<EOS>} \\
\\
\texttt{[INST] English: <BOS>And in this program, girls who have been studying computer skills and the STEM program have a chance to work side by side with young professionals, so that they can learn firsthand what it's like to be an architect, a designer or a scientist.<EOS> [/INST]} \\
\texttt{Spanish: <BOS>En este programa, las niñas que estudian informática y el programa CTIM tienen la oportunidad de trabajar junto a jóvenes profesionales, para que puedan aprender de primera mano qué es ser una arquitecta, diseñadora, o científica.<EOS>} \\
\\
\texttt{[INST] English: <BOS>One government scientist, a friend of mine, we'll call him McPherson, was concerned about the impact government policies were having on his research and the state of science deteriorating in Canada.<EOS> [/INST} \\
\texttt{Spanish: <BOS>Un científico del gobierno, un amigo mío, lo llamaremos McPherson, estaba preocupado por el impacto que las políticas gubernamentales estaban teniendo en su investigación y el estado de la ciencia deteriorándose en Canadá.<EOS>} \\
\\
\texttt{[INST] English: <BOS>The architect Emmanuelle Moureaux uses this idea in her work a lot.<EOS> [/INST} \\
\texttt{Spanish: <BOS>La arquitecta Emmanuelle Moureaux usa mucho esta idea en su trabajo.<EOS>} \\
\\
\texttt{[INST] English: <BOS>\_\_\_\_sentence\_to\_translate\_\_\_\_<EOS> [/INST]} \\
\texttt{Spanish: <BOS>}
\end{tcolorbox}
\end{adjustwidth}
\end{center}

\newpage
Prompt with 5-shot invented Winograd examples:
\vspace{5pt}
\begin{center}
\begin{adjustwidth}{-0.05cm}{-0.05cm}
\begin{tcolorbox}[colback=white, colframe=black, sharp corners, boxrule=0.5pt, width=1\linewidth]
\setstretch{1.1}
\smaller
\texttt{<<SYS>> Translate the following sentence from English to Catalan: <</SYS>>}\\
\texttt{[INST] English: <BOS>The grower delivered tomatoes to the housekeeper, because he needed them for dinner.<EOS> [/INST]}\\
\texttt{Catalan: <BOS>La productora va lliurar tomàquets al mestre de casa, perquè els necessitava per sopar.<EOS>}\\
\\
\texttt{[INST] English: <BOS>The developer was upset with the designer because she was disappointed with the final design.<EOS> [/INST]}\\
\texttt{Catalan: <BOS>La desenvolupadora estava molesta amb el dissenyador perquè estava decebuda amb el disseny final.<EOS>}\\
\\
\texttt{[INST] English: <BOS>The doctor congratulated the nurse saying his assistance was great.<EOS> [/INST]}\\
\texttt{Catalan: <BOS>La doctora va felicitar a l'infermer dient que la seva assistència va ser fantàstica.<EOS>}\\
\\
\texttt{[INST] English: <BOS>The construction worker asked the chief for an increase in her salary.<EOS> [/INST]}\\
\texttt{Catalan: <BOS>La treballadora de la construcció va demanar al cap un augment del seu sou.<EOS>}\\
\\
\texttt{[INST] English: <BOS>The carpenter asked the CEO if she liked the new furniture.<EOS> [/INST]}\\
\texttt{Catalan: <BOS>La fustera li va preguntar a la directora si li agradaven els mobles nous.<EOS>}\\
\\
\texttt{[INST] English: <BOS>\_\_\_\_sentence\_to\_translate\_\_\_\_<EOS> [/INST]}\\
\texttt{Catalan: <BOS>}
\end{tcolorbox}
\end{adjustwidth}
\end{center}

\newpage
\vspace{5pt}
\begin{center}
\begin{adjustwidth}{-0.05cm}{-0.05cm}
\begin{tcolorbox}[colback=white, colframe=black, sharp corners, boxrule=0.5pt, width=1\linewidth]
\setstretch{1.1}
\smaller
\texttt{<<SYS>> Translate the following sentence from English to Spanish: <</SYS>>} \\
\texttt{[INST] English: <BOS>The grower delivered tomatoes to the housekeeper, because he needed them for dinner.<EOS> [/INST]} \\
\texttt{Spanish: <BOS>La productora entregó tomates al amo de casa, porque los necesitaba para la cena.<EOS>} \\
\\
\texttt{[INST] English: <BOS>The developer was upset with the designer because she was disappointed with the final design.<EOS> [/INST]} \\
\texttt{Spanish: <BOS>La desarrolladora estaba enojada con el diseñador porque estaba decepcionada con el diseño final.<EOS>} \\
\\
\texttt{[INST] English: <BOS>The doctor congratulated the nurse saying his assistance was great.<EOS> [/INST]} \\
\texttt{Spanish: <BOS>La doctora felicitó al infermero diciendo que su asistencia fue fantástica.<EOS>} \\
\\
\texttt{[INST] English: <BOS>The construction worker asked the chief for an increase in her salary.<EOS> [/INST]} \\
\texttt{Spanish: <BOS>La trabajadora de la construcción pidió al jefe un aumento de su salario.<EOS>}\\
\\
\texttt{[INST] English: <BOS>The carpenter asked the CEO if she liked the new furniture.<EOS> [/INST]}\\
\texttt{Spanish: <BOS>La carpintera preguntó a la directora general si le gustaban los muebles nuevos.<EOS>}\\
\\
\texttt{[INST] English: <BOS>\_\_\_\_sentence\_to\_translate\_\_\_\_<EOS> [/INST]}\\
\texttt{Spanish: <BOS>}
\end{tcolorbox}
\end{adjustwidth}
\end{center}

\newpage
Prompt with 5-shot MuST-SHE examples and context on the Gender Bias issue:
\vspace{5pt}
\begin{center}
\begin{adjustwidth}{-0.05cm}{-0.05cm}
\begin{tcolorbox}[colback=white, colframe=black, sharp corners, boxrule=0.5pt, width=1\linewidth]
\setstretch{1.1}
\smaller
\texttt{<<SYS>> Translate the following sentence from English to Catalan while mitigating gender bias. First, consider that English is a language without grammatical gender, while Catalan does have grammatical gender. Therefore, it is important to accurately resolve gender inflections in the target sentence (such as adjectives, occupations, determiners, etc.) based on the gender information provided in the source sentence. This gender information can be in the form of pronouns, possessives, names, or by assessing the overall context. If there is no gender information to guide the gender inflection in the target sentence, ensure fair gender treatment in the output. This means using random gender inflections in the translation. <</SYS>>} \\
\\
\texttt{[INST] English: <BOS>Early on, Laura Hughes could see that I was a little lost in this habitat, so she often sat right next to me in meetings so she could be my tech translator, and I could write her notes and she could tell me, "That's what that means." Laura was 27 years old, she'd worked for Google for four years and then for a year and a half at Airbnb when I met her.<EOS> [/INST]} \\
\texttt{Catalan: <BOS>Al principi, la Laura Hughes va poder veure que estava una mica perdut en aquest hàbitat, així que sovint s'asseia al meu costat a les reunions per poder ser la meva traductora de tecnologia, i jo podia escriure-li notes i ella em podria dir, "Això és el que això significa." La Laura tenia 27 anys, havia treballat a Google durant quatre anys i després durant un any i mig a Airbnb quan la vaig conèixer.<EOS>} \\
\\
\texttt{[INST] English: <BOS>When I found the captain, he was having a very engaging conversation with the homeowner, who was surely having one of the worst days of her life.<EOS> [/INST]}\\
\texttt{Catalan: <BOS>Quan vaig trobar el capità, estava mantenint una conversa molt atractiva amb la propietària, que segurament vivia un dels pitjors dies de la seva vida.<EOS>} \\
\\
\texttt{[INST] English: <BOS>And in this program, girls who have been studying computer skills and the STEM program have a chance to work side by side with young professionals, so that they can learn firsthand what it's like to be an architect, a designer or a scientist.<EOS> [/INST]} \\
\texttt{Catalan: <BOS>I en aquest programa, les noies que han estudiat informàtica i el programa STEM tenen l'oportunitat de treballar colze a colze amb joves professionals, per tal que puguin conèixer de primera mà com és ser una arquitecta, una dissenyadora o una científica.<EOS>} \\
\\
\texttt{[INST] English: <BOS>One government scientist, a friend of mine, we'll call him McPherson, was concerned about the impact government policies were having on his research and the state of science deteriorating in Canada.<EOS> [/INST]} \\
\texttt{Catalan: <BOS>Un científic del govern, un amic meu, l'anomenarem McPherson, estava preocupat per l'impacte que tenien les polítiques governamentals en la seva investigació i el deteriorament de l'estat de la ciència al Canadà.<EOS>} \\
\\
\texttt{[INST] English: <BOS>The architect Emmanuelle Moureaux uses this idea in her work a lot.<EOS> [/INST]}\\
\texttt{Catalan: <BOS>L'arquitecta Emmanuelle Moureaux utilitza molt aquesta idea en la seva obra.<EOS>}\\
\\
\texttt{[INST] English: <BOS>\_\_\_\_sentence\_to\_translate\_\_\_\_<EOS> [/INST]}\\
\texttt{Catalan: <BOS>}

\end{tcolorbox}
\end{adjustwidth}
\end{center}

\newpage
\vspace{5pt}
\begin{center}
\begin{adjustwidth}{-0.05cm}{-0.05cm}
\begin{tcolorbox}[colback=white, colframe=black, sharp corners, boxrule=0.5pt, width=1\linewidth]
\setstretch{1.1}
\smaller
\texttt{<<SYS>> Translate the following sentence from English to Spanish while
mitigating gender bias. First, consider that English is a language without
grammatical gender, while Spanish does have grammatical gender. Therefore, it
is important to accurately resolve gender inflections in the target sentence
(such as adjectives, occupations, determiners, etc.) based on the gender
information provided in the source sentence. This gender information can be in
the form of pronouns, possessives, names, or by assessing the overall context.
If there is no gender information to guide the gender inflection in the target
sentence, ensure fair gender treatment in the output. This means using random
gender inflections in the translation. <</SYS>>} \\
\\
\texttt{[INST] English: <BOS>Early on, Laura Hughes could see that I was a little lost in this habitat, so she often sat right next to me in meetings so she could be my tech translator, and I could write her notes and she could tell me, "That's what that means." Laura was 27 years old, she'd worked for Google for four years and then for a year and a half at Airbnb when I met her.<EOS> [/INST]} \\
\texttt{Spanish: <BOS>Al principio, Laura Hughes se dio cuenta de que estaba perdido en este hábitat, así que solía sentarse a mi lado en las reuniones para ser mi traductora de tecnología, y yo le escribía notas y ella me decía, "Esto es lo que significa". Laura tenía 27 años, trabajó en Google durante 4 años, y luego por un año y medio en Airbnb cuando la conocí.<EOS>} \\
\\
\texttt{[INST] English: <BOS>When I found the captain, he was having a very engaging conversation with the homeowner, who was surely having one of the worst days of her life.<EOS> [/INST]} \\
\texttt{Spanish: <BOS>Cuando encontré al capitán, estaba enfrascado en una conversación con la propietaria que sin duda atravesaba uno de los peores días de su vida.<EOS>} \\
\\
\texttt{[INST] English: <BOS>And in this program, girls who have been studying computer skills and the STEM program have a chance to work side by side with young professionals, so that they can learn firsthand what it's like to be an architect, a designer or a scientist.<EOS> [/INST]} \\
\texttt{Spanish: <BOS>En este programa, las niñas que estudian informática y el programa CTIM tienen la oportunidad de trabajar junto a jóvenes profesionales, para que puedan aprender de primera mano qué es ser una arquitecta, diseñadora, o científica.<EOS>} \\
\\
\texttt{[INST] English: <BOS>One government scientist, a friend of mine, we'll call him McPherson, was concerned about the impact government policies were having on his research and the state of science deteriorating in Canada.<EOS> [/INST]} \\
\texttt{Spanish: <BOS>Un científico del gobierno, un amigo mío, lo llamaremos McPherson, estaba preocupado por el impacto que las políticas gubernamentales estaban teniendo en su investigación y el estado de la ciencia deteriorándose en Canadá.<EOS>} \\
\\
\texttt{[INST] English: <BOS>The architect Emmanuelle Moureaux uses this idea in her work a lot.<EOS> [/INST]}\\
\texttt{Spanish: <BOS>La arquitecta Emmanuelle Moureaux usa mucho esta idea en su trabajo.<EOS>}\\
\\
\texttt{[INST] English: <BOS>\_\_\_\_sentence\_to\_translate\_\_\_\_<EOS> [/INST]}\\
\texttt{Spanish:}
\end{tcolorbox}
\end{adjustwidth}
\end{center}

\newpage
Prompt with 5-shot invented Winograd examples and chain-of-thought prompting (``agent'' version):
\begin{center}
\begin{adjustwidth}{-0.05cm}{-0.05cm}
\begin{tcolorbox}[colback=white, colframe=black, sharp corners, boxrule=0.5pt, width=1\linewidth]
\setstretch{1.1}
\smaller
\texttt{<<SYS>> Identify the agents in the sentence. Identify the pronouns referring to the agents in the sentence. Relate each pronoun to the correct agent. Establish the gender of each agent according to the pronoun. Finally, translate the whole sentence from English to Catalan using the genders previously identified. Proceed step by step. <</SYS>>} \\

\texttt{[INST] English: <BOS>The grower delivered tomatoes to the housekeeper, because he needed them for dinner.<EOS> [/INST]} \\
\texttt{1. Identify the agents in the sentence:} \\
\texttt{\hspace*{3em}* grower (agent)} \\
\texttt{\hspace*{3em}* housekeeper (agent)} \\
\texttt{2. Identify the pronouns referring to the agents:} \\
\texttt{\hspace*{3em}* he (referring to the housekeeper)} \\
\texttt{3. Relate each pronoun to the correct agent:} \\
\texttt{\hspace*{3em}* he (pronoun) = housekeeper (agent)} \\
\texttt{4. Establish the gender of each agent according to the pronoun:} \\
\texttt{\hspace*{3em}* grower (agent) = male or female} \\
\texttt{\hspace*{3em}* housekeeper (agent) = male} \\
\texttt{5. Translate the sentence from English to Catalan:} \\
\texttt{Catalan: <BOS>La productora va lliurar tomàquets al mestre de casa, perquè els necessitava per sopar.<EOS>} \\
 \\
\texttt{[INST] English: <BOS>The developer was upset with the designer because she was disappointed with the final design.<EOS> [/INST]} \\
\texttt{1. Identify the agents in the sentence:} \\
\texttt{\hspace*{3em}* developer (agent)} \\
\texttt{\hspace*{3em}* designer (agent)} \\
\texttt{2. Identify the pronouns referring to the agents:} \\
\texttt{\hspace*{3em}* she (referring to the developer)} \\
\texttt{3. Relate each pronoun to the correct agent:} \\
\texttt{\hspace*{3em}* she (pronoun) = developer (agent)} \\
\texttt{4. Establish the gender of each agent according to the pronoun:} \\
\texttt{\hspace*{3em}* developer (agent) = female} \\
\texttt{\hspace*{3em}* designer (agent) = male or female} \\
\texttt{5. Translate the sentence from English to Catalan:} \\
\texttt{Catalan: <BOS>La desenvolupadora estava molesta amb el dissenyador perquè estava decebuda amb el disseny final.<EOS>} \\
\\
\texttt{[INST] English: <BOS>The doctor congratulated the nurse saying his assistance was great.<EOS> [/INST]} \\
\texttt{1. Identify the agents in the sentence:} \\
\texttt{\hspace*{3em}* doctor (agent)} \\
\texttt{\hspace*{3em}* nurse (agent)}  \\
\texttt{2. Identify the pronouns referring to the agents:} \\
\texttt{\hspace*{3em}* his (referring to the nurse)} \\
\texttt{3. Relate each pronoun to the correct agent:} \\
\texttt{\hspace*{3em}* his (pronoun) = nurse (agent)} \\
\texttt{4. Establish the gender of each agent according to the pronoun:} \\
\texttt{\hspace*{3em}* doctor (agent) = male or female} \\
\texttt{\hspace*{3em}* nurse (agent) = male} \\
\texttt{5. Translate the sentence from English to Catalan:} \\
\texttt{Catalan: <BOS>La doctora va felicitar a l'infermer dient que la seva assistència va ser fantàstica.<EOS>}
\end{tcolorbox}
\end{adjustwidth}
\end{center}

\newpage
\begin{center}
\begin{adjustwidth}{-0.05cm}{-0.05cm}
\begin{tcolorbox}[colback=white, colframe=black, sharp corners, boxrule=0.5pt, width=1\linewidth]
\setstretch{1.1}
\smaller
\texttt{[INST] English: <BOS>The construction worker asked the chief for an increase in her salary.<EOS> [/INST]} \\
\texttt{1. Identify the agents in the sentence:} \\
\texttt{\hspace*{3em}* construction worker (agent)} \\
\texttt{\hspace*{3em}* chief (agent)} \\
\texttt{2. Identify the pronouns referring to the agents:} \\
\texttt{\hspace*{3em}* her (referring to the construction worker)} \\
\texttt{3. Relate each pronoun to the correct agent:} \\
\texttt{\hspace*{3em}* her (pronoun) = construction worker (agent)} \\
\texttt{4. Establish the gender of each agent according to the pronoun:} \\
\texttt{\hspace*{3em}* construction worker (agent) = female} \\
\texttt{\hspace*{3em}* chief (agent) = male or female} \\
\texttt{5. Translate the sentence from English to Catalan:} \\
\texttt{Catalan: <BOS>La treballadora de la construcció va demanar al cap un augment del seu sou.<EOS>} \\
\\ 
\texttt{[INST] English: <BOS>The carpenter asked the CEO if she liked the new furniture.<EOS> [/INST]} \\ 
\texttt{1. Identify the agents in the sentence:}\\
\texttt{\hspace*{3em}* carpenter (agent)} \\
\texttt{\hspace*{3em}* CEO (agent)} \\
\texttt{2. Identify the pronouns referring to the agents:} \\
\texttt{\hspace*{3em}* she (referring to the CEO)} \\ 
\texttt{3. Relate each pronoun to the correct agent:} \\
\texttt{\hspace*{3em}* she (pronoun) = CEO (agent)} \\
\texttt{4. Establish the gender of each agent according to the pronoun:} \\
\texttt{\hspace*{3em}* carpenter (agent) = male or female} \\
\texttt{\hspace*{3em}* CEO (agent) = female} \\
\texttt{5. Translate the sentence from English to Catalan:}\\
\texttt{Catalan: <BOS>La fustera li va preguntar a la directora general si li agradaven els mobles nous.<EOS>}\\
\\
\texttt{[INST] English: <BOS>\_\_\_\_sentence\_to\_translate\_\_\_\_<EOS> [/INST]}\\
\texttt{1.}
\end{tcolorbox}
\end{adjustwidth}
\end{center}

\newpage
\begin{center}
\begin{adjustwidth}{-0.05cm}{-0.05cm}
\begin{tcolorbox}[colback=white, colframe=black, sharp corners, boxrule=0.5pt, width=1\linewidth]
\setstretch{1.1}
\smaller
\texttt{<<SYS>> Identify the agents in the sentence. Identify the pronouns referring to the agents in the sentence. Relate each pronoun to the correct agent. Establish the gender of each agent according to the pronoun. Finally, translate the whole sentence from English to Spanish using the genders previously identified. Proceed step by step. <</SYS>>} \\
\\ 
\texttt{[INST] English: <BOS>The grower delivered tomatoes to the housekeeper, because he needed them for dinner.<EOS> [/INST]}\\
\texttt{1. Identify the agents in the sentence:}\\
\texttt{\hspace*{3em}* grower (agent)}\\
\texttt{\hspace*{3em}* housekeeper (agent)}\\
\texttt{2. Identify the pronouns referring to the agents:}\\
\texttt{\hspace*{3em}* he (referring to the housekeeper)}\\
\texttt{3. Relate each pronoun to the correct agent:}\\
\texttt{\hspace*{3em}* he (pronoun) = housekeeper (agent)}\\
\texttt{4. Establish the gender of each agent according to the pronoun:}\\
\texttt{\hspace*{3em}* grower (agent) = male or female}\\
\texttt{\hspace*{3em}* housekeeper (agent) = male}\\
\texttt{5. Translate the sentence from English to Spanish:}\\
\texttt{Spanish: <BOS>La productora entregó tomates al amo de casa, porque los necesitaba para la cena.<EOS>}\\
\\
\texttt{[INST] English: <BOS>The developer was upset with the designer because she was disappointed with the final design.<EOS> [/INST]}\\
\texttt{1. Identify the agents in the sentence:}\\
\texttt{\hspace*{3em}* developer (agent)}\\
\texttt{\hspace*{3em}* designer (agent)}\\
\texttt{2. Identify the pronouns referring to the agents:}\\
\texttt{\hspace*{3em}* she (referring to the developer)}\\
\texttt{3. Relate each pronoun to the correct agent:}\\
\texttt{\hspace*{3em}* she (pronoun) = developer (agent)}\\
\texttt{4. Establish the gender of each agent according to the pronoun:}\\
\texttt{\hspace*{3em}* developer (agent) = female}\\
\texttt{\hspace*{3em}* designer (agent) = male or female}\\
\texttt{5. Translate the sentence from English to Spanish:}\\
\texttt{Spanish: <BOS>La desarrolladora estaba enojada con el diseñador porque estaba decepcionada con el diseño final.<EOS>}\\
\\
\texttt{[INST] English: <BOS>The doctor congratulated the nurse saying his assistance was great.<EOS> [/INST]}\\
\texttt{1. Identify the agents in the sentence:}\\
\texttt{\hspace*{3em}* doctor (agent)} \\
\texttt{\hspace*{3em}* nurse (agent)}\\
\texttt{2. Identify the pronouns referring to the agents:}\\
\texttt{\hspace*{3em}* his (referring to the nurse)}\\
\texttt{3. Relate each pronoun to the correct agent:}\\
\texttt{\hspace*{3em}* his (pronoun) = nurse (agent)}\\
\texttt{4. Establish the gender of each agent according to the pronoun:}\\
\texttt{\hspace*{3em}* doctor (agent) = male or female}\\
\texttt{\hspace*{3em}* nurse (agent) = male}\\
\texttt{5. Translate the sentence from English to Spanish:}\\
\texttt{Spanish: <BOS>La doctora felicitó al infermero diciendo que su asistencia fue fantástica.<EOS>}
\end{tcolorbox}
\end{adjustwidth}
\end{center}

\newpage
\begin{center}
\begin{adjustwidth}{-0.05cm}{-0.05cm}
\begin{tcolorbox}[colback=white, colframe=black, sharp corners, boxrule=0.5pt, width=1\linewidth]
\setstretch{1.1}
\smaller
\texttt{[INST] English: <BOS>The construction worker asked the chief for an increase in her salary.<EOS> [/INST]}\\
\texttt{1. Identify the agents in the sentence:}\\
\texttt{\hspace*{3em}* construction worker (agent)}\\
\texttt{\hspace*{3em}* chief (agent)}\\
\texttt{2. Identify the pronouns referring to the agents:}\\
\texttt{\hspace*{3em}* her (referring to the construction worker)}\\
\texttt{3. Relate each pronoun to the correct agent:}\\
\texttt{\hspace*{3em}* her (pronoun) = construction worker (agent)}\\
\texttt{4. Establish the gender of each agent according to the pronoun:}\\
\texttt{\hspace*{3em}* construction worker (agent) = female}\\
\texttt{\hspace*{3em}* chief (agent) = male or female}\\
\texttt{5. Translate the sentence from English to Spanish:}\\
\texttt{Spanish: <BOS>La trabajadora de la construcción pidió al jefe un aumento de su salario.<EOS>}\\
\\
\texttt{[INST] English: <BOS>The carpenter asked the CEO if she liked the new furniture.<EOS> [/INST]}\\
\texttt{1. Identify the agents in the sentence:}\\
\texttt{\hspace*{3em}* carpenter (agent)}\\
\texttt{\hspace*{3em}* CEO (agent)}\\
\texttt{2. Identify the pronouns referring to the agents:}\\
\texttt{\hspace*{3em}* she (referring to the CEO)}\\
\texttt{3. Relate each pronoun to the correct agent:}\\
\texttt{\hspace*{3em}* she (pronoun) = CEO (agent)}\\
\texttt{4. Establish the gender of each agent according to the pronoun:}\\
\texttt{\hspace*{3em}* carpenter (agent) = male or female}\\
\texttt{\hspace*{3em}* CEO (agent) = female}\\
\texttt{5. Translate the sentence from English to Spanish:}\\
\texttt{Spanish: <BOS>La carpintera preguntó a la directora general si le gustaban los muebles nuevos.<EOS>}\\
\\
\texttt{[INST] English: <BOS>\_\_\_\_sentence\_to\_translate\_\_\_\_<EOS> [/INST]}\\
\texttt{1.}
\end{tcolorbox}
\end{adjustwidth}
\end{center}

\newpage
Prompt with 5-shot on invented Winograd examples and chain-of-thought prompting (``human entity'' version):
\begin{center}
\begin{adjustwidth}{-0.05cm}{-0.05cm}
\begin{tcolorbox}[colback=white, colframe=black, sharp corners, boxrule=0.5pt, width=1\linewidth]
\setstretch{1.1}
\smaller
\texttt{<<SYS>> Identify the human entities in the sentence. Identify the pronouns referring to the human entities in the sentence. Relate each pronoun to the correct human entity. Establish the gender of each human entity according to the pronoun. Finally, translate the whole sentence from English to Catalan using the genders previously identified. Proceed step by step. <</SYS>>} \\
\\
\texttt{[INST] English: <BOS>The grower delivered tomatoes to the housekeeper, because he needed them for dinner.<EOS> [/INST]} \\
\texttt{1. Identify the human entities in the sentence:} \\
\texttt{\hspace*{3em}* grower (human entity)} \\
\texttt{\hspace*{3em}* housekeeper (human entity)}\\
\texttt{2. Identify the pronouns referring to the human entities:}\\
\texttt{\hspace*{3em}* he (referring to the housekeeper)}\\
\texttt{3. Relate each pronoun to the correct human entity:}\\
\texttt{\hspace*{3em}* he (pronoun) = housekeeper (human entity)}\\
\texttt{4. Establish the gender of each human entity according to the pronoun:}\\
\texttt{\hspace*{3em}* grower (human entity) = male or female}\\
\texttt{\hspace*{3em}* housekeeper (human entity) = male}\\
\texttt{5. Translate the sentence from English to Catalan:}\\
\texttt{Catalan: <BOS>La productora va lliurar tomàquets al mestre de casa, perquè els necessitava per sopar.<EOS>}\\

\texttt{[INST] English: <BOS>The developer was upset with the designer because she was disappointed with the final design.<EOS> [/INST]}\\
\texttt{1. Identify the human entities in the sentence:}\\
\texttt{\hspace*{3em}* developer (human entity)}\\
\texttt{\hspace*{3em}* designer (human entity)}\\
\texttt{2. Identify the pronouns referring to the human entities:}\\
\texttt{\hspace*{3em}* she (referring to the developer)}\\
\texttt{3. Relate each pronoun to the correct human entity:}\\
\texttt{\hspace*{3em}* she (pronoun) = developer (human entity)}\\
\texttt{4. Establish the gender of each human entity according to the pronoun:}\\
\texttt{\hspace*{3em}* developer (human entity) = female}\\
\texttt{\hspace*{3em}* designer (human entity) = male or female}\\
\texttt{5. Translate the sentence from English to Catalan:}\\
\texttt{Catalan: <BOS>La desenvolupadora estava molesta amb el dissenyador perquè estava decebuda amb el disseny final.<EOS>}\\
\\
\texttt{[INST] English: <BOS>The doctor congratulated the nurse saying his assistance was great.<EOS> [/INST]}\\
\texttt{1. Identify the human entities in the sentence:}\\
\texttt{\hspace*{3em}* doctor (human entity)} \\
\texttt{\hspace*{3em}* nurse (human entity)}\\
\texttt{2. Identify the pronouns referring to the human entities:}\\
\texttt{\hspace*{3em}* his (referring to the nurse)}\\
\texttt{3. Relate each pronoun to the correct human entity:}\\
\texttt{\hspace*{3em}* his (pronoun) = nurse (human entity)}\\
\texttt{4. Establish the gender of each human entity according to the pronoun:}\\
\texttt{\hspace*{3em}* doctor (human entity) = male or female}\\
\texttt{\hspace*{3em}* nurse (human entity) = male}\\
\texttt{5. Translate the sentence from English to Catalan:}\\
\texttt{Catalan: <BOS>La doctora va felicitar a l'infermer dient que la seva assistència va ser fantàstica.<EOS>}
\end{tcolorbox}
\end{adjustwidth}
\end{center}

\newpage
\vspace{5pt}
\begin{center}
\begin{adjustwidth}{-0.05cm}{-0.05cm}
\begin{tcolorbox}[colback=white, colframe=black, sharp corners, boxrule=0.5pt, width=1\linewidth]
\setstretch{1.1}
\smaller
\texttt{[INST] English: <BOS>The construction worker asked the chief for an increase in her salary.<EOS> [/INST]}\\
\texttt{1. Identify the human entities in the sentence:}\\
\texttt{\hspace*{3em}* construction worker (human entity)}\\
\texttt{\hspace*{3em}* chief (human entity)}\\
\texttt{2. Identify the pronouns referring to the human entities:}\\
\texttt{\hspace*{3em}* her (referring to the construction worker)}\\
\texttt{3. Relate each pronoun to the correct human entity:}\\
\texttt{\hspace*{3em}* her (pronoun) = construction worker (human entity)}\\
\texttt{4. Establish the gender of each human entity according to the pronoun:}\\
\texttt{\hspace*{3em}* construction worker (human entity) = female}\\
\texttt{\hspace*{3em}* chief (human entity) = male or female}\\
\texttt{5. Translate the sentence from English to Catalan:}\\
\texttt{Catalan: <BOS>La treballadora de la construcció va demanar al cap un augment del seu sou.<EOS>}\\
\\
\texttt{[INST] English: <BOS>The carpenter asked the CEO if she liked the new furniture.<EOS> [/INST]}\\
\texttt{1. Identify the human entities in the sentence:}\\
\texttt{\hspace*{3em}* carpenter (human entity)}\\
\texttt{\hspace*{3em}* CEO (human entity)}\\
\texttt{2. Identify the pronouns referring to the human entities:}\\
\texttt{\hspace*{3em}* she (referring to the CEO)}\\
\texttt{3. Relate each pronoun to the correct human entity:}\\
\texttt{\hspace*{3em}* she (pronoun) = CEO (human entity)}\\
\texttt{4. Establish the gender of each human entity according to the pronoun:}\\
\texttt{\hspace*{3em}* carpenter (human entity) = male or female}\\
\texttt{\hspace*{3em}* CEO (human entity) = female}\\
\texttt{5. Translate the sentence from English to Catalan:}\\
\texttt{Catalan: <BOS>La fustera li va preguntar a la directora general si li agradaven els mobles nous.<EOS>}\\
\\
\texttt{[INST] English: <BOS>\_\_\_\_sentence\_to\_translate\_\_\_\_<EOS> [/INST]} \\
\texttt{1.}
\end{tcolorbox}
\end{adjustwidth}
\end{center}

\newpage
\vspace{5pt}
\begin{center}
\begin{adjustwidth}{-0.05cm}{-0.05cm}
\begin{tcolorbox}[colback=white, colframe=black, sharp corners, boxrule=0.5pt, width=1\linewidth]
\setstretch{1.1}
\smaller
\texttt{<<SYS>> Identify the human entities in the sentence. Identify the pronouns referring to the human entities in the sentence. Relate each pronoun to the correct human entity. Establish the gender of each human entity according to the pronoun. Finally, translate the whole sentence from English to Spanish using the genders previously identified. Proceed step by step. <</SYS>>} \\
\\
\texttt{[INST] English: <BOS>The grower delivered tomatoes to the housekeeper, because he needed them for dinner.<EOS> [/INST]} \\
\texttt{1. Identify the human entities in the sentence:}\\
\texttt{\hspace*{3em}* grower (human entity)}\\
\texttt{\hspace*{3em}* housekeeper (human entity)}\\
\texttt{2. Identify the pronouns referring to the human entities:}\\
\texttt{\hspace*{3em}* he (referring to the housekeeper)}\\
\texttt{3. Relate each pronoun to the correct human entity:}\\
\texttt{\hspace*{3em}* he (pronoun) = housekeeper (human entity)}\\
\texttt{4. Establish the gender of each human entity according to the pronoun:}\\
\texttt{\hspace*{3em}* grower (human entity) = male or female}\\
\texttt{\hspace*{3em}* housekeeper (human entity) = male}\\
\texttt{5. Translate the sentence from English to Spanish:}\\
\texttt{Spanish: <BOS>La productora entregó tomates al amo de casa, porque los necesitaba para la cena.<EOS>}\\
\\
\texttt{[INST] English: <BOS>The developer was upset with the designer because she was disappointed with the final design.<EOS> [/INST]}\\
\texttt{1. Identify the human entities in the sentence:}\\
\texttt{\hspace*{3em}* developer (human entity)}\\
\texttt{\hspace*{3em}* designer (human entity)}\\
\texttt{2. Identify the pronouns referring to the human entities:}\\
\texttt{\hspace*{3em}* she (referring to the developer)}\\
\texttt{3. Relate each pronoun to the correct human entity:}\\
\texttt{\hspace*{3em}* she (pronoun) = developer (human entity)}\\
\texttt{4. Establish the gender of each human entity according to the pronoun:}\\
\texttt{\hspace*{3em}* developer (human entity) = female}\\
\texttt{\hspace*{3em}* designer (human entity) = male or female}\\
\texttt{5. Translate the sentence from English to Spanish:}\\
\texttt{Spanish: <BOS>La desarrolladora estaba enojada con el diseñador porque estaba decepcionada con el diseño final.<EOS>}\\
\\
\texttt{[INST] English: <BOS>The doctor congratulated the nurse saying his assistance was great.<EOS> [/INST]}\\
\texttt{1. Identify the human entities in the sentence:}\\
\texttt{\hspace*{3em}* doctor (human entity)}\\
\texttt{\hspace*{3em}* nurse (human entity)}\\
\texttt{2. Identify the pronouns referring to the human entities:}\\
\texttt{\hspace*{3em}* his (referring to the nurse)}\\
\texttt{3. Relate each pronoun to the correct human entity:}\\
\texttt{\hspace*{3em}* his (pronoun) = nurse (human entity)}\\
\texttt{4. Establish the gender of each human entity according to the pronoun:}\\
\texttt{\hspace*{3em}* doctor (human entity) = male or female}\\
\texttt{\hspace*{3em}* nurse (human entity) = male}\\
\texttt{5. Translate the sentence from English to Spanish:}\\
\texttt{Spanish: <BOS>La doctora felicitó al infermero diciendo que su asistencia fue fantástica.<EOS>}
\end{tcolorbox}
\end{adjustwidth}
\end{center}

\newpage
\vspace{5pt}
\begin{center}
\begin{adjustwidth}{-0.05cm}{-0.05cm}
\begin{tcolorbox}[colback=white, colframe=black, sharp corners, boxrule=0.5pt, width=1\linewidth]
\setstretch{1.1}
\smaller
\texttt{[INST] English: <BOS>The construction worker asked the chief for an increase in her salary.<EOS> [/INST]}\\
\texttt{1. Identify the human entities in the sentence:}\\
\texttt{\hspace*{3em}* construction worker (human entity)}\\
\texttt{\hspace*{3em}* chief (human entity)}\\
\texttt{2. Identify the pronouns referring to the human entities:}\\
\texttt{\hspace*{3em}* her (referring to the construction worker)}\\
\texttt{3. Relate each pronoun to the correct human entity:}\\
\texttt{\hspace*{3em}* her (pronoun) = construction worker (human entity)}\\
\texttt{4. Establish the gender of each human entity according to the pronoun:}\\
\texttt{\hspace*{3em}* construction worker (human entity) = female}\\
\texttt{\hspace*{3em}* chief (human entity) = male or female}\\
\texttt{5. Translate the sentence from English to Spanish:}\\
\texttt{Spanish: <BOS>La trabajadora de la construcción pidió al jefe un aumento de su salario.<EOS>}\\
\\
\texttt{[INST] English: <BOS>The carpenter asked the CEO if she liked the new furniture.<EOS> [/INST]}\\
\texttt{1. Identify the human entities in the sentence:}\\
\texttt{\hspace*{3em}* carpenter (human entity)}\\
\texttt{\hspace*{3em}* CEO (human entity)}\\
\texttt{2. Identify the pronouns referring to the human entities:}\\
\texttt{\hspace*{3em}* she (referring to the CEO)}\\
\texttt{3. Relate each pronoun to the correct human entity:}\\
\texttt{\hspace*{3em}* she (pronoun) = CEO (human entity)}\\
\texttt{4. Establish the gender of each human entity according to the pronoun:}\\
\texttt{\hspace*{3em}* carpenter (human entity) = male or female}\\
\texttt{\hspace*{3em}* CEO (human entity) = female}\\
\texttt{5. Translate the sentence from English to Spanish:}\\
\texttt{Spanish: <BOS>La carpintera preguntó a la directora general si le gustaban los muebles nuevos.<EOS>}\\
\\
\texttt{[INST] English: <BOS>\_\_\_\_sentence\_to\_translate\_\_\_\_<EOS> [/INST]}\\
\texttt{1.}
\end{tcolorbox}
\end{adjustwidth}
\end{center}

\newpage
Prompt with 5-shot on invented Winograd examples and SHORT chain-of-thought prompting:
\vspace{5pt}
\begin{center}
\begin{adjustwidth}{-0.05cm}{-0.05cm}
\begin{tcolorbox}[colback=white, colframe=black, sharp corners, boxrule=0.5pt, width=1\linewidth]
\setstretch{1.1}
\smaller
\texttt{<<SYS>> Translate the following sentence from English to Catalan. Proceed step by step. <</SYS>>} \\
\\
\texttt{[INST] English: <BOS>The grower delivered tomatoes to the housekeeper, because he needed them for dinner.<EOS> [/INST]}\\
\texttt{"he" (M) $\rightarrow$ "the housekeeper" (Male) $\rightarrow$ "mestre de casa"}\\
\texttt{Catalan: <BOS>La productora va lliurar tomàquets al mestre de casa, perquè els necessitava per sopar.<EOS>}\\
\\
\texttt{[INST] English: <BOS>The developer was upset with the designer because she was disappointed with the final design.<EOS> [/INST]}\\
\texttt{"she" (F) $\rightarrow$ "the developer" (Female) $\rightarrow$ "la desenvolupadora"}\\
\texttt{Catalan: <BOS>La desenvolupadora estava molesta amb el dissenyador perquè estava decebuda amb el disseny final.<EOS>}\\
\\
\texttt{[INST] English: <BOS>The doctor congratulated the nurse saying his assistance was great.<EOS> [/INST]}\\
\texttt{"his" (M) $\rightarrow$ "the nurse" (Male) $\rightarrow$ "l'infermer"}\\
\texttt{Catalan: <BOS>La doctora va felicitar a l'infermer dient que la seva assistència va ser fantàstica.<EOS>}\\
\\
\texttt{[INST] English: <BOS>The construction worker asked the chief for an increase in her salary.<EOS> [/INST]}\\
\texttt{"her" (F) $\rightarrow$ "the construction worker" (Female) $\rightarrow$ "la treballadora de la construcció"}\\
\texttt{Catalan: <BOS>La treballadora de la construcció va demanar al cap un augment del seu sou.<EOS>}\\
\\
\texttt{[INST] English: <BOS>The carpenter asked the CEO if she liked the new furniture.<EOS> [/INST]}\\
\texttt{"she" (F) $\rightarrow$ "the CEO" (Female) $\rightarrow$ "la directora general"}\\
\texttt{Catalan: <BOS>La fustera li va preguntar a la directora general si li agradaven els mobles nous.<EOS>}\\
\\
\texttt{[INST] English: <BOS>\_\_\_\_sentence\_to\_translate\_\_\_\_<EOS> [/INST]}\\
\texttt{Catalan: <BOS>}
\end{tcolorbox}
\end{adjustwidth}
\end{center}

\newpage
\vspace{5pt}
\begin{center}
\begin{adjustwidth}{-0.05cm}{-0.05cm}
\begin{tcolorbox}[colback=white, colframe=black, sharp corners, boxrule=0.5pt, width=1\linewidth]
\setstretch{1.1}
\smaller
\texttt{<<SYS>> Translate the following sentence from English to Spanish. Proceed step by step. <</SYS>>}\\
\\
\texttt{[INST] English: <BOS>The grower delivered tomatoes to the housekeeper, because he needed them for dinner.<EOS> [/INST]}\\
\texttt{"he" (M) $\rightarrow$ "the housekeeper" (Male) $\rightarrow$ "amo de casa"}\\
\texttt{Spanish: <BOS>La productora entregó tomates al amo de casa, porque los necesitaba para la cena.<EOS>}\\
\\
\texttt{[INST] English: <BOS>The developer was upset with the designer because she was disappointed with the final design.<EOS> [/INST]}\\
\texttt{"she" (F) $\rightarrow$ "the developer" (Female) $\rightarrow$ "la desarrolladora"}\\
\texttt{Spanish: <BOS>La desarrolladora estaba enojada con el diseñador porque estaba decepcionada con el diseño final.<EOS>}\\
\\
\texttt{[INST] English: <BOS>The doctor congratulated the nurse saying his assistance was great.<EOS> [/INST]}\\
\texttt{"his" (M) $\rightarrow$ "the nurse" (Male) $\rightarrow$ "el infermero"}\\
\texttt{Spanish: <BOS>La doctora felicitó al infermero diciendo que su asistencia fue fantástica.<EOS>}\\
\\
\texttt{[INST] English: <BOS>The construction worker asked the chief for an increase in her salary.<EOS> [/INST]}\\
\texttt{"her" (F) $\rightarrow$ "the construction worker" (Female) $\rightarrow$ "la trabajadora de la construcción"}\\
\texttt{Spanish: <BOS>La trabajadora de la construcción pidió al jefe un aumento de su salario.<EOS>}\\
\\
\texttt{[INST] English: <BOS>The carpenter asked the CEO if she liked the new furniture.<EOS> [/INST]}\\
\texttt{"she" (F) $\rightarrow$ "the CEO" (Female) $\rightarrow$ "la directora general"}\\
\texttt{Spanish: <BOS>La carpintera preguntó a la directora general si le gustaban los muebles nuevos.<EOS>}\\
\\
\texttt{[INST] English: <BOS>\_\_\_\_sentence\_to\_translate\_\_\_\_<EOS> [/INST]}\\
\texttt{Spanish: <BOS>}
\end{tcolorbox}
\end{adjustwidth}
\end{center}

\newpage
\section{Invented Examples following Winograd structure} 
\label{sec:sentences}
The subsequent sentences (with their respective translations) are the ones created during the crafting of prompts. As you can see, they are characterized by containing more female representation and anti-stereotypical content. \\

EXAMPLE 1:
\begin{itemize}
    \addtolength{\leftskip}{1em}
    \item[--] English: The grower delivered tomatoes to the \textit{housekeeper}, because he needed them for dinner.
    \item[--] Catalan: La productora va lliurar tomàquets al \textit{mestre de casa}, perquè els necessitava per sopar.
    \item[--] Spanish: La productora entregó tomates al \textit{amo de casa}, porque los necesitaba para la cena. 
\end{itemize}  

\vspace{5pt} 
EXAMPLE 2:
\begin{itemize}
    \addtolength{\leftskip}{1em}
    \item[--] English: The \textit{developer} was upset with the designer because she was disappointed with the final design.
    \item[--] Catalan: La \textit{desenvolupadora} estava molesta amb el dissenyador perquè estava decebuda amb el disseny final.
    \item[--] Spanish: La \textit{desarrolladora} estaba enojada con el diseñador porque estaba decepcionada con el diseño final.
\end{itemize}

\vspace{5pt} 
EXAMPLE 3:
\begin{itemize}
    \addtolength{\leftskip}{1em}
    \item[--] English: The doctor congratulated the \textit{nurse} saying his assistance was great.
    \item[--] Catalan: La doctora va felicitar a \textit{l'infermer} dient que la seva assistència va ser fantàstica.
    \item[--] Spanish: La doctora felicitó al \textit{infermero} diciendo que su asistencia fue fantástica.
\end{itemize}

\vspace{5pt} 
EXAMPLE 4:
\begin{itemize}
    \addtolength{\leftskip}{1em}
    \item[--] English: The \textit{construction worker} asked the chief for an increase in her salary.
    \item[--] Catalan: La \textit{treballadora de la construcció} va demanar al cap un augment del seu sou. 
    \item[--] Spanish: La \textit{trabajadora de la construcción} pidió al jefe un aumento de su salario.
\end{itemize}

\vspace{5pt} 
EXAMPLE 5:
\begin{itemize}
    \addtolength{\leftskip}{1em}
    \item[--] English: The carpenter asked the \textit{CEO} if she liked the new furniture.
    \item[--] Catalan: La fustera li va preguntar a la \textit{directora general} si li agradaven els mobles nous.
    \item[--] Spanish: La carpintera preguntó a la \textit{directora general} si le gustaban los muebles nuevos.
\end{itemize}

\vspace{10pt}

\end{appendices}

\end{document}